\documentclass{article}
\usepackage[T1]{fontenc}
\usepackage{microtype}
\usepackage{graphicx}
\usepackage{subcaption}
\usepackage{booktabs}
\usepackage[dvipsnames]{xcolor}
\usepackage{hyperref}
\usepackage[preprint]{icml2026}
\setcitestyle{numbers,square,comma,sort&compress}
\usepackage{amsmath,amssymb,amsthm}
\usepackage{xspace}
\usepackage[capitalize,noabbrev]{cleveref}
\usepackage{cuted}
\usepackage{placeins}
\usepackage{multirow}
\usepackage{colortbl}
\definecolor{vaportint}{RGB}{253,239,236}
\newtheorem{proposition}{Proposition}

\makeatletter
\renewcommand{\ICML@preprint}{\textit{Preprint.}}
\makeatother
\hypersetup{
    pdftitle={VAPOR: Improving Multimodal Reasoning with Sparse Visual Updates},
    pdfauthor={Mingyu Su, Wei Wu, Erle Zhu, Chengxing Xie, Deyao Hong, Hongning Wang},
    pdfsubject={Preprint},
    pdfkeywords={Multimodal reasoning, efficient inference, visual attention}
}
\icmltitlerunning{VAPOR: Multimodal Reasoning with Sparse Visual Updates}

\begin{document}
\twocolumn[
\icmltitle{VAPOR: Improving Multimodal Reasoning with Sparse Visual Updates}
\begin{icmlauthorlist}
    \icmlauthor{Mingyu Su}{coai}
    \icmlauthor{Wei Wu}{ant}
    \icmlauthor{Erle Zhu}{coai}
    \icmlauthor{Chengxing Xie}{coai}
    \icmlauthor{Deyao Hong}{coai}
    \icmlauthor{Hongning Wang}{coai}
\par\smallskip
    {\small\normalfont
    \href{mailto:sumy25@mails.tsinghua.edu.cn}{\texttt{sumy25@mails.tsinghua.edu.cn}}\quad
    \href{mailto:hw-ai@tsinghua.edu.cn}{\texttt{hw-ai@tsinghua.edu.cn}}}
\end{icmlauthorlist}
\icmlaffiliation{coai}{The Conversational AI (CoAI) Group, Tsinghua University}
\icmlaffiliation{ant}{Ant Group}
\icmlcorrespondingauthor{Hongning Wang}{hw-ai@tsinghua.edu.cn}

\vskip 0.15in
]
\printAffiliationsAndNotice{}
\begin{strip}
    \centering
    \includegraphics[width=\linewidth]{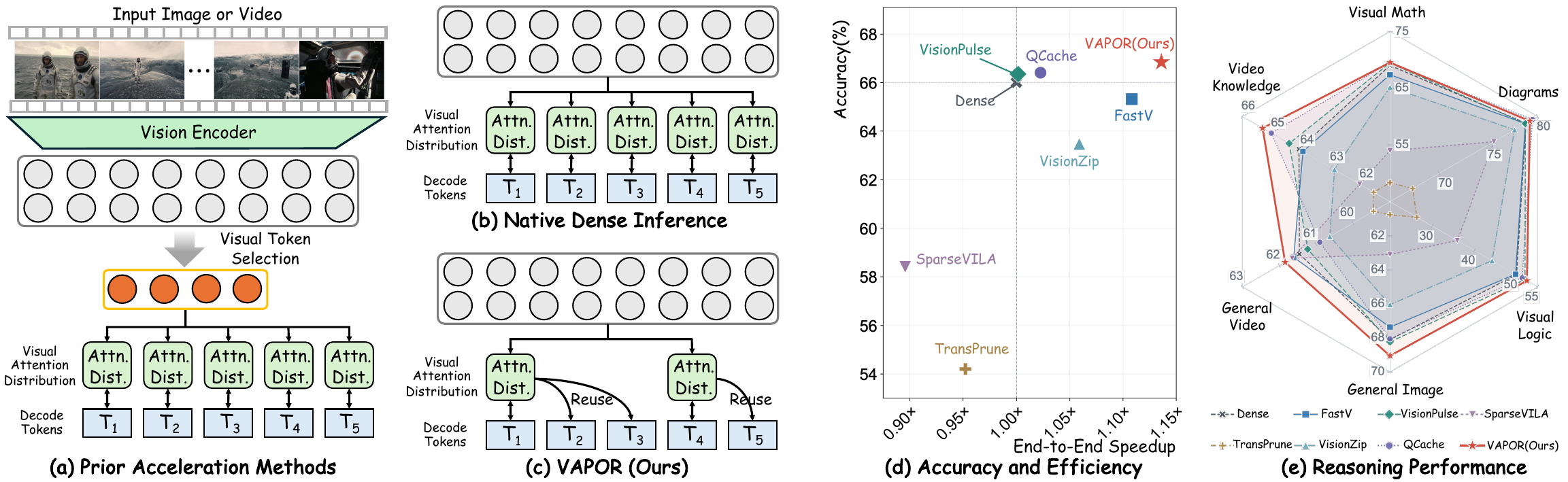}
    \captionsetup{hypcap=false}
    \captionof{figure}{
        \textbf{VAPOR decouples visual updates from token-by-token decoding.}
        (a) Token-selection-based acceleration reduces the visual tokens attended to at each decoding step.
        (b) Dense inference recomputes attention over all visual tokens at every step.
        (c) VAPOR reuses cached visual attention outputs between sparse, adaptive refreshes, while textual attention updates at every step.
        (d) VAPOR improves both accuracy and inference speed over the compared methods.
        (e) Accuracy across six task categories.
        Results use Qwen3-VL-8B on eight image reasoning and four video reasoning benchmarks.
        Speedup is measured relative to Dense using end-to-end inference time.
    }
    \label{fig:intro}
    \captionsetup{hypcap=true}
\end{strip}
\begin{abstract}
Multimodal large language models (MLLMs) have made substantial progress in visual reasoning, but extended generation over large visual contexts remains computationally expensive. During autoregressive decoding, visual attention is recomputed for every generated token, even when consecutive tokens rely on similar visual evidence. We introduce \textbf{VAPOR} (Visual Attention Persistence with On-demand Refresh), a training-free framework that reduces this redundancy by reusing visual attention outputs across decoding steps and refreshing them when needed. VAPOR caches the visual attention output while continuing to update textual attention at every step, helping reduce visual grounding drift during extended reasoning. To decide when to reuse cached outputs, VAPOR estimates visual attention mass using a low-rank approximation. It also uses prefill-based attention consistency and cache age to make independent Persist and Refresh decisions for each attention head. VAPOR retains the full visual key--value (KV) cache, so all visual tokens remain available when new evidence is needed. Experiments on eight image and four video reasoning benchmarks across two MLLM families show that VAPOR improves accuracy over dense attention on nearly all benchmarks while achieving 1.11--1.21$\times$ decoding speedups. Compared with the strongest state-of-the-art acceleration baseline in each setting, VAPOR further improves accuracy by 0.35--2.15 percentage points while achieving comparable or lower decoding latency.
Our code is available at \url{https://github.com/Su-my/VAPOR}.
\end{abstract}

\section{Introduction}
\label{sec:introduction}

Multimodal large language models (MLLMs) have advanced image and video understanding and increasingly support multi-step reasoning over visual inputs~\cite{team2026qwen3,team2026gemma,clark2026molmo2,yang2025r1,feng2026video}. Yet these capabilities come at a cost: high-resolution images and long videos produce large visual contexts, and extended reasoning repeatedly invokes attention over them~\cite{chen2024image,zhang2024sparsevlm,xing2024pyramiddrop,yang2025visionzip}. In standard autoregressive decoding, each generated token induces a new attention query, even when successive tokens concern the same visual observation. The model thus recomputes visual attention during passages that may not require new visual evidence. This raises a simple question: \emph{must visual attention be recomputed at every decoding step?}

Existing acceleration techniques prune or merge visual tokens~\cite{chen2024image,shang2025llava,yang2025visionzip,zhang2024sparsevlm}, dynamically select visual information during decoding~\cite{khaki2025sparsevila,xu2026visionpulse}, or reuse attention across layers~\cite{zhuang2026q,bhojanapalli2021leveraging,xiao2019sharing}. We investigate a complementary choice: \emph{when should visual attention outputs be reused, and when should they be refreshed?} Consecutive tokens may play different linguistic roles while drawing on the same visual evidence. Recomputing the visual component at every step can therefore waste computation and introduce variations in visual attention as the textual context evolves. Conversely, keeping that component fixed indefinitely risks missing evidence required by later reasoning steps. An effective approach must preserve useful visual information while detecting when a fresh visual attention computation is warranted.

We propose \textbf{VAPOR} (\textbf{V}isual \textbf{A}ttention \textbf{P}ersistence with \textbf{O}n-demand \textbf{R}efresh), a training-free framework for adaptive reuse of visual attention outputs during decoding (\cref{fig:intro}). VAPOR decomposes attention into a conditional visual output, its visual mixing weight, and a textual attention output. Each head alternates between \emph{Persist}, which reuses its cached conditional visual output, and \emph{Refresh}, which recomputes exact attention and updates the cache. A lightweight low-rank estimator of visual attention mass guides these decisions. The refresh threshold is calibrated using prefill consistency and adjusted according to cache age to avoid indefinite reuse. Textual attention continues to update at every step, while the full visual key--value (KV) cache remains available for subsequent \emph{Refresh} steps. This allows VAPOR to reduce repeated visual attention computation without permanently pruning visual tokens.

We evaluate VAPOR on eight image and four video reasoning benchmarks across two MLLM families.
Compared with Dense, VAPOR improves average accuracy by 0.43--0.85 percentage points and achieves 1.11--1.21$\times$ decoding speedups, with higher accuracy on nearly all model--benchmark comparisons.
It also outperforms the strongest evaluated acceleration baseline in average accuracy across all four model--modality settings, with gains of 0.35--2.15 percentage points and comparable or lower decoding latency.
Ablation studies further validate the adaptive refresh design by examining threshold and cache-age sensitivity, prefill-based calibration, and proxy-rank effects.
A qualitative case study illustrates how visual attention persistence helps maintain focus on task-relevant visual evidence during reasoning.

Our contributions are summarized as follows:
\begin{itemize}
    \item We introduce \textbf{VAPOR}, a training-free framework for adaptive visual attention reuse. It decouples visual attention refresh from token-by-token generation through head-wise \emph{Persist} and \emph{Refresh} decisions, while retaining the full visual KV cache.

    \item We develop an efficient visual-mass estimator based on low-rank approximation and an adaptive refresh mechanism that combines prefill-based attention consistency with cache age, enabling selective reuse of visual attention outputs.

    \item We conduct extensive evaluations on eight image and four video reasoning benchmarks across two MLLM families. Comprehensive ablation studies and analyses provide insights into the effectiveness and behavior of VAPOR.
\end{itemize}

\begin{figure*}[t]
    \centering
    \includegraphics[width=\textwidth]{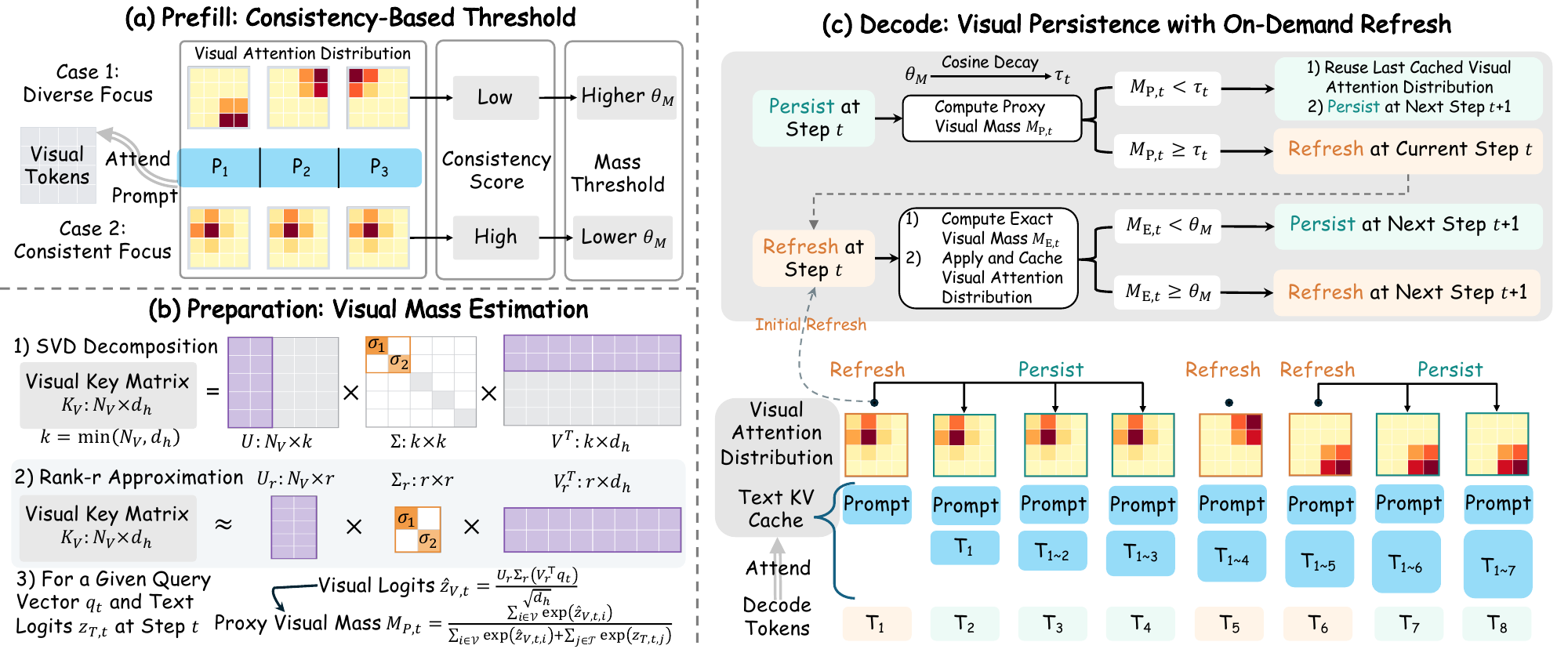}
    \caption{
        \textbf{Overview of VAPOR.}
        (a) Prefill attention consistency determines a base visual-mass threshold $\theta_M$.
        (b) A low-rank approximation of the visual keys provides low-cost estimates of visual logits and proxy visual mass $M_{P,t}$ during decoding.
        (c) Each attention head independently switches between \emph{Refresh}, which computes and caches the exact visual attention output, and \emph{Persist}, which reuses the cached output.
        During Persist, proxy mass above an age-adjusted threshold triggers an immediate Refresh. After Refresh, exact visual mass $M_{E,t}$ determines the next state.
        The full visual KV cache is retained, and textual attention is updated at every step.
    }
    \label{fig:overview}
\end{figure*}

\section{Related Work}
\label{sec:related_work}

\smallskip\noindent\textbf{Multimodal Reasoning.}
Building on the language understanding and generation capabilities of pretrained LLMs~\cite{brown2020language, zhang2022opt, touvron2023llama, yang2025qwen3, liu2025deepseek}, MLLMs extend language modeling to visual inputs.
Early approaches connect pretrained vision encoders with LLMs through cross-modal alignment, while visual instruction tuning enables models to follow image-grounded instructions~\cite{alayrac2022flamingo, li2023blip, liu2023visual}.
Recent models have advanced high-resolution image and video understanding~\cite{bai2025qwen3, team2026qwen3, team2026gemma, clark2026molmo2}.
Meanwhile, reasoning-oriented training has further enhanced their ability to perform multi-step reasoning over visual inputs~\cite{yang2025r1, feng2026video}.
However, high-resolution images and long videos introduce large numbers of visual tokens, increasing the cost of attention computation during extended reasoning~\cite{chen2024image, yang2025visionzip}.
These advances make efficient processing of visual information increasingly important.

\smallskip\noindent\textbf{Efficient Multimodal Inference.}
Existing methods reduce multimodal inference costs through visual token compression or attention reuse.
Visual representations can be compressed using learned bottlenecks, pooling, or attention distillation~\cite{li2023blip, alayrac2022flamingo, yao2024deco, ye2025voco}.
Other approaches reduce visual tokens through selection, merging, and attention-guided pruning~\cite{shang2025llava, yang2025visionzip, chen2024image, zhang2024sparsevlm, xing2024pyramiddrop}.
More recent methods dynamically select visual tokens during decoding to adapt to changing information needs~\cite{khaki2025sparsevila, xu2026visionpulse}.
However, reducing the visual tokens accessed at each step may exclude fine-grained details needed for subsequent reasoning~\cite{endo2025feather}.

Attention reuse provides an alternative to token reduction.
Although KV caching avoids recomputing past keys and values, attention must still be evaluated for each new query.
Prior work exploits cross-layer attention similarity to share or reuse attention scores and weights, including in MLLM decoding~\cite{bhojanapalli2021leveraging, xiao2019sharing, zhuang2026q}.
In contrast, VAPOR reuses visual attention outputs across decoding steps while retaining the full visual KV cache and refreshing attention when new evidence is needed.

\smallskip\noindent\textbf{Language Priors and Visual Grounding.}
Pretrained language priors provide useful knowledge for multimodal reasoning but can also bias models toward plausible responses that conflict with visual evidence~\cite{li2023evaluating, guan2024hallusionbench, favero2024multi}.
Recent studies show that extended reasoning can weaken visual grounding as models increasingly rely on linguistic context rather than visual inputs~\cite{xu2026more, jin2026look, peng2026deeper, luo2026thinking}.
Existing mitigation methods include preference-based training~\cite{zhou2024aligning, zhao2023beyond, deng2026learning}, contrastive decoding~\cite{leng2024mitigating, lee2025retrieval}, and visual attention interventions~\cite{peng2026deeper}.
VAPOR complements these approaches by adaptively reusing and refreshing visual attention outputs, reducing visual grounding drift during extended reasoning without permanently discarding visual information.

\section{Method}
\label{sec:method}
\label{sec:vapor}

VAPOR decouples visual attention refresh from token generation through independent head-wise decisions (\cref{fig:overview}). It estimates visual attention mass at each decoding step and reuses the cached conditional visual output unless demand or cache age calls for a refresh. Textual attention continues to update, and the full visual key--value (KV) cache remains available. We introduce the attention decomposition and its empirical motivation, then describe the reuse mechanism and its computational cost.

\begin{figure*}[t]
    \centering
    \includegraphics[width=\textwidth]{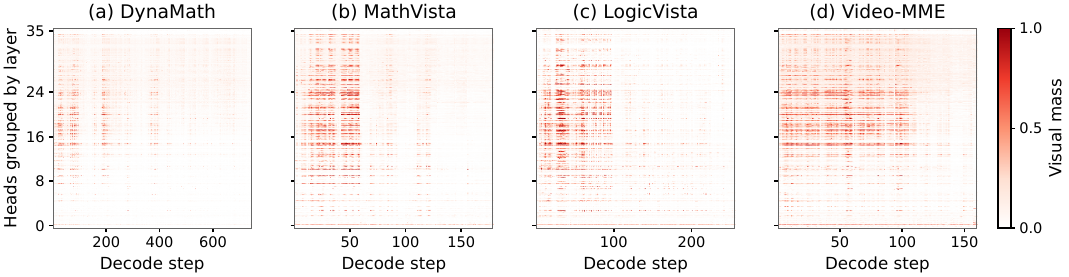}
    \caption{
        \textbf{Visual attention mass during decoding.}
        Each panel shows a randomly selected Dense decoding trajectory from Qwen3-VL-8B-Instruct on one of four benchmarks: (a) DynaMath, (b) MathVista, (c) LogicVista, and (d) Video-MME.
        Rows denote attention heads grouped by layer, columns denote decoding steps, and color indicates exact visual attention mass (white: low; red: high).
    }
    \label{fig:motivation}
\end{figure*}

\subsection{Attention Decomposition and Motivation}
\label{sec:attention_output}
\label{sec:motivation}

For a decoding query $q_t\in\mathbb{R}^{d_h}$, let $\mathcal V$ and $\mathcal T_t$ denote the visual and nonvisual keys, respectively. We omit layer and head indices, as all quantities and decisions are computed independently per head. Using position-encoded queries and keys, define the attention logits and partition sums as
\begin{equation}
\begin{aligned}
    z_{t,i} = \frac{q_t^{\mathsf T}k_i}{\sqrt{d_h}},\quad
    Z_{V,t} = \sum_{i\in\mathcal V}e^{z_{t,i}},\quad
    Z_{T,t} = \sum_{j\in\mathcal T_t}e^{z_{t,j}}.
\end{aligned}
\end{equation}
The \emph{exact visual mass} is $M_{E,t}=Z_{V,t}/(Z_{V,t}+Z_{T,t})$. The corresponding conditional visual and textual attention outputs are
\begin{equation}
\begin{aligned}
    \bar v_t &= \sum_{i\in\mathcal V}\frac{e^{z_{t,i}}}{Z_{V,t}}v_i, &
    \bar u_t &= \sum_{j\in\mathcal T_t}\frac{e^{z_{t,j}}}{Z_{T,t}}v_j.
\end{aligned}
\end{equation}
The full attention output then admits the exact decomposition
\begin{equation}
    o_t=M_{E,t}\bar v_t+(1-M_{E,t})\bar u_t.
    \label{eq:attention_decomposition}
\end{equation}
This decomposition separates visual attention mass from the visual content it aggregates. Since visual values remain fixed during decoding, the conditional output $\bar v_t$ can be cached without storing the full attention distribution.

\smallskip\noindent\textbf{Empirical motivation.}
\Cref{fig:motivation} reveals two patterns in four randomly selected Dense decoding trajectories from Qwen3-VL-8B-Instruct. First, high visual mass is sparse across \emph{head--step pairs}: only 0.34\%--2.58\% satisfy $M_{E,t}\geq0.5$. These high-mass steps also vary across heads, motivating independent refresh decisions. Second, visual mass generally declines over decoding, with 90.0\%--95.9\% of heads showing a lower mean mass in the final quarter than in the first.

These patterns motivate selective, head-wise reuse. From \cref{eq:attention_decomposition}, replacing $\bar v_t$ with a cached output $\bar v_a$ changes the full attention output by $M_{E,t}(\bar v_a-\bar v_t)$ when the visual mass is exact and the textual component is fixed. Low visual mass therefore attenuates reuse error, though it does not make visual information irrelevant. Reuse may also reduce query-induced fluctuations in visual routing. The challenge is to estimate visual demand without computing the full visual attention.

\subsection{Adaptive Visual Attention Reuse}
\label{sec:persistence_framework}

VAPOR combines low-cost visual-mass estimation with head-wise threshold control. Each head alternates between \emph{Persist}, which reuses its cached conditional visual output, and \emph{Refresh}, which recomputes exact visual attention and updates the cache.

\subsubsection{Efficient Visual Mass Estimation}
\label{sec:visual_mass}

After prefill, we construct a rank-$r$ approximation of each distinct cached visual key matrix $K_V\in\mathbb{R}^{N_V\times d_h}$:
\begin{equation}
    K_V\approx U_r\Sigma_rV_r^{\mathsf T},
\end{equation}
where $U_r$, $\Sigma_r$, and $V_r$ contain the leading singular components and $r\ll\min(N_V,d_h)$. Computed once after prefill, the decomposition is shared by query heads using the same visual keys. For each query, the estimated visual logits are
\begin{equation}
    \hat z_{V,t}=\frac{U_r\Sigma_r(V_r^{\mathsf T}q_t)}{\sqrt{d_h}}.
    \label{eq:proxy_logits}
\end{equation}
Absorbing $\Sigma_r$ into $U_r$ reduces the per-query cost from $O(N_Vd_h)$ to $O(rd_h+N_Vr)$. All visual positions are retained; only the projection dimension is reduced.

We compute text logits exactly and combine their partition sum with the estimated visual logits to obtain the \emph{proxy visual mass}:
\begin{equation}
    M_{P,t}=\frac{\sum_{i\in\mathcal V}e^{\hat z_{V,t,i}}}
    {\sum_{i\in\mathcal V}e^{\hat z_{V,t,i}}+Z_{T,t}}.
    \label{eq:proxy_mass}
\end{equation}
The proxy requires no visual value aggregation. During Persist, it controls refresh decisions and supplies the visual mixing weight, while the conditional visual output remains cached from the last exact computation.

\begin{proposition}[Visual-mass approximation]
\label{prop:mass_bound}
For a fixed head and query $q_t$, let $\widehat K_V=U_r\Sigma_rV_r^{\mathsf T}$ be the truncated SVD of $K_V$.
With the same exact text partition sum, in exact arithmetic,
\begin{equation}
    |M_{P,t}-M_{E,t}|
    \leq \frac{\sigma_{r+1}(K_V)\|q_t\|_2}{4\sqrt{d_h}}
    =:b_{r,t},
    \label{eq:mass_error_bound}
\end{equation}
where $\sigma_{r+1}(K_V)$ is the largest discarded singular value.
\end{proposition}
Thus, during Persist, a margin $|M_{E,t}-\tau_{h,t}|>b_{r,t}$ guarantees that proxy and exact mass yield the same threshold decision. The proof and a tighter bound are given in Sec.~\ref{app:mass_bound}.

\subsubsection{Head-wise Persist and Refresh}
\label{sec:refresh}

Each head maintains a \emph{Persist}/\emph{Refresh} state, an anchor step $a_h$, and a cached conditional visual output $\bar v_{a_h}$. The first decoding step performs Refresh to initialize the cache. Refresh computes exact attention over all available keys and updates the cached output. If $M_{E,t}<\theta_M$, the head enters Persist at the next step. Otherwise, it remains in Refresh.

During Persist, the head evaluates the proxy visual mass $M_{P,t}$. If it falls below the current reuse threshold $\tau_{h,t}$, defined in \cref{sec:consistency}, the head reuses its cached conditional visual output and returns
\begin{equation}
    \hat o_t=M_{P,t}\bar v_{a_h}+(1-M_{P,t})\bar u_t.
    \label{eq:persist_output}
\end{equation}
If $M_{P,t}\geq\tau_{h,t}$, the head refreshes immediately, so the triggering token receives exact visual attention. The resulting exact mass determines the next state, and the maximum cache age can also force Refresh.

During Persist, $\bar u_t$ and $M_{P,t}$ are updated at every step, while the conditional visual output remains fixed at $\bar v_{a_h}$. This preserves current textual attention and allows independent refresh decisions across heads. The full visual KV cache remains available for every Refresh.

\subsubsection{Adaptive Threshold Control}
\label{sec:consistency}

The refresh policy calibrates a base threshold from prefill consistency and adjusts it according to each head's cache age.

\smallskip\noindent\textbf{Prefill-based calibration.}
Let $\pi^{V,h}_u$ denote head $h$'s conditional visual attention distribution for prompt token $u$ after the visual input. We compute the average Bhattacharyya overlap over all pairs of $n\geq2$ valid prompt tokens:
\begin{equation}
    C_h=\frac{2}{n(n-1)}\sum_{u<v}\sum_{i\in\mathcal V}
    \sqrt{\pi^{V,h}_{u,i}\pi^{V,h}_{v,i}}.
\end{equation}
We take the median of $C_h$ across valid layer--head pairs as the input-level consistency score $C$ and set the base threshold to $\theta_M=f(C)$ using a predetermined non-increasing mapping. This score captures agreement across prompt tokens, not temporal stability during decoding. We use it as a heuristic cue for visual routing: higher consistency gives a lower refresh threshold, while lower consistency requires a stronger visual-mass signal before replacing the cached output. The base threshold is shared across heads, but refresh decisions remain head-specific.

\smallskip\noindent\textbf{Age-based adjustment.}
We define cache age as $\delta_{h,t}=t-a_h$, where $a_h$ is the most recent anchor step. For a maximum age $D>2$, the Persist threshold is
\begin{equation}
\begin{aligned}
    s(\delta) = \operatorname{clip}\!\left(\frac{\delta-2}{D-2},0,1\right),\quad
    \tau_{h,t} = \theta_M\cos\!\left(\frac{\pi}{2}s(\delta_{h,t})\right).
\end{aligned}
\label{eq:age_threshold}
\end{equation}
The threshold remains at $\theta_M$ for the first two steps after an anchor and then decreases with cache age, making Refresh increasingly likely. At age $D$, Refresh is forced and the head's age is reset. Because the schedule depends on time since the last refresh rather than absolute decoding position, each head is controlled independently.

\subsection{Efficiency Analysis}
\label{sec:complexity}

We analyze decoding-attention matrix products. Let $H$ be the total number of query heads, $T$ the decoding length, and $\alpha$ the fraction of head--step pairs that perform Refresh. Exact visual query--key and attention--value products cost $4N_Vd_h$ FLOPs per pair. Including nonvisual keys, Dense requires
\begin{equation}
    F_{\mathrm{Dense}}=4Hd_h\sum_{t=1}^{T}(N_V+|\mathcal T_t|).
    \label{eq:dense_flops}
\end{equation}

During Persist, VAPOR replaces the exact visual products with a rank-$r$ proxy costing $2r(N_V+d_h)$ FLOPs. Textual attention remains exact. Let $\beta$ be the fraction of head--step pairs that evaluate the proxy, including those that trigger Refresh, so $1-\alpha\leq\beta\leq1$. Reusing text logits for a triggered Refresh leaves textual cost unchanged. The saved computation, $\Delta F=F_{\mathrm{Dense}}-F_{\mathrm{VAPOR}}$, is
\begin{equation}
    \Delta F=2HT\bigl[2(1-\alpha)N_Vd_h-\beta r(N_V+d_h)\bigr].
    \label{eq:attention_flops}
\end{equation}
Setting $\beta=1$ gives a conservative lower bound for a given refresh fraction. This analysis counts decoding attention matrix products only, excluding elementwise operations and one-time preparation costs.

\begin{table*}[t]
    \centering
    \caption{\textbf{Accuracy and mean decoding latency on image and video reasoning benchmarks.}
        Acc and Time denote accuracy (\%, $\uparrow$) and latency (s/sample, $\downarrow$), respectively.
        Avg is the unweighted mean across benchmarks within each modality.
        Best and second-best results within each model are marked in \textbf{bold} and \underline{underline}, respectively.}
    \label{tab:main_results}
    \begingroup
    \footnotesize
    \setlength{\tabcolsep}{1pt}
    \renewcommand{\arraystretch}{1.25}
    \begin{tabular}{@{}cll*{14}{c}@{}}
        \toprule
        Model & Method & & \multicolumn{9}{c}{\textbf{Image Reasoning}} & \multicolumn{5}{c}{\textbf{Video Reasoning}} \\
        \cmidrule(lr){4-12} \cmidrule(lr){13-17}
         & & & \shortstack{Math\\Vista} & \shortstack{Math\\Verse} & \shortstack{Dyna\\Math} & WeMath & AI2D & \shortstack{Logic\\Vista} & MMBench & \shortstack{MMMU\\Pro} & \textbf{Avg} & MVBench & \shortstack{Video\\MME} & MMVU & \shortstack{Video\\MMMU} & \textbf{Avg} \\
        \midrule
         &  & Acc~$\uparrow$ & 74.90 & \underline{62.39} & 65.85 & 72.87 & 78.82 & 50.00 & 86.22 & 49.97 & 67.63 & \underline{62.80} & 60.11 & 66.72 & 61.44 & 62.77 \\
         & \multirow{-2}{*}{Dense} & Time~$\downarrow$ & 53.2 & 102.1 & 74.0 & 102.4 & 47.8 & 160.9 & 18.7 & 137.4 & 87.1 & 17.6 & 30.3 & 25.5 & 75.0 & 37.1 \\
        \cline{2-17}
         &  & Acc~$\uparrow$ & 72.70 & 61.60 & 62.67 & 72.47 & 78.76 & 49.78 & 85.54 & 49.20 & 66.59 & 62.55 & \underline{60.67} & 66.56 & 61.33 & 62.78 \\
         & \multirow{-2}{*}{FastV} & Time~$\downarrow$ & \underline{50.8} & \textbf{88.3} & \underline{68.1} & \underline{90.5} & \textbf{40.2} & \underline{151.1} & 19.1 & \underline{127.9} & \underline{79.5} & 16.5 & \textbf{25.4} & 21.4 & 71.8 & 33.8 \\
        \cline{2-17}
         &  & Acc~$\uparrow$ & \textbf{76.20} & 62.16 & 66.39 & \underline{72.99} & 78.76 & \underline{52.01} & \underline{86.31} & \underline{50.15} & 68.12 & 62.05 & 60.41 & 66.72 & 62.11 & 62.82 \\
         & \multirow{-2}{*}{VisionPulse} & Time~$\downarrow$ & 54.6 & 103.7 & 72.8 & 101.9 & 48.8 & 161.1 & 19.0 & 137.1 & 87.4 & 16.6 & 28.7 & 22.6 & 77.1 & 36.3 \\
        \cline{2-17}
         &  & Acc~$\uparrow$ & 54.70 & 50.63 & 43.23 & 67.30 & 75.55 & 35.94 & 84.97 & 41.22 & 56.69 & 62.45 & \textbf{60.85} & 65.60 & 58.44 & 61.84 \\
         & \multirow{-2}{*}{SparseVILA} & Time~$\downarrow$ & 83.5 & 114.6 & 96.5 & 101.4 & 45.6 & 186.5 & 19.4 & 149.9 & 99.7 & \textbf{16.0} & \underline{25.7} & \underline{19.4} & 76.8 & 34.5 \\
        \cline{2-17}
         &  & Acc~$\uparrow$ & 48.10 & 47.16 & 36.51 & 61.55 & 67.29 & 26.34 & 83.16 & 38.36 & 51.06 & 59.48 & 59.41 & 65.76 & 57.33 & 60.49 \\
         & \multirow{-2}{*}{TransPrune} & Time~$\downarrow$ & 68.1 & 105.8 & 81.5 & 107.7 & 45.8 & 179.8 & 20.5 & 149.2 & 94.8 & 19.0 & 26.0 & 20.5 & 72.7 & 34.5 \\
        \cline{2-17}
         &  & Acc~$\uparrow$ & 70.50 & 59.70 & 60.76 & 69.71 & 77.66 & 44.20 & 85.74 & 46.34 & 64.32 & 61.85 & 59.41 & \underline{67.20} & 58.56 & 61.75 \\
         & \multirow{-2}{*}{VisionZip} & Time~$\downarrow$ & 53.2 & 96.6 & 72.2 & 93.3 & 42.9 & 163.5 & \textbf{16.2} & 137.3 & 84.4 & \underline{16.3} & 26.3 & \textbf{15.7} & 73.5 & \underline{32.9} \\
        \cline{2-17}
         &  & Acc~$\uparrow$ & 75.00 & \textbf{62.69} & \textbf{67.01} & \textbf{73.45} & \textbf{79.47} & 51.34 & 86.09 & 50.03 & \underline{68.13} & 61.50 & 60.30 & 67.04 & \underline{63.00} & \underline{62.96} \\
         & \multirow{-2}{*}{QCache} & Time~$\downarrow$ & 53.6 & 99.2 & 69.9 & 101.1 & 46.9 & 161.0 & 18.9 & 138.5 & 86.1 & 18.6 & 29.0 & 23.4 & \underline{68.9} & 35.0 \\
        \cline{2-17}
        \rowcolor{vaportint}
        \cellcolor{white} &  & Acc~$\uparrow$ & \underline{76.00} & \textbf{62.69} & \underline{66.45} & 72.93 & \underline{79.21} & \textbf{52.46} & \textbf{86.52} & \textbf{51.58} & \textbf{68.48} & \textbf{63.20} & 60.48 & \textbf{67.52} & \textbf{63.11} & \textbf{63.58} \\
        \rowcolor{vaportint}
        \cellcolor{white}\multirow{-16}{*}{\rotatebox{90}{\textbf{Qwen3-VL-8B}}} & \multirow{-2}{*}{\textbf{VAPOR}} & Time~$\downarrow$ & \textbf{47.8} & \underline{90.4} & \textbf{64.1} & \textbf{89.1} & \underline{42.6} & \textbf{135.1} & \underline{16.5} & \textbf{119.8} & \textbf{75.7} & 16.5 & \textbf{25.4} & 22.1 & \textbf{66.5} & \textbf{32.6} \\
        \midrule
         &  & Acc~$\uparrow$ & \underline{73.30} & \underline{54.67} & \underline{62.93} & \underline{53.91} & 82.51 & \underline{50.00} & \textbf{83.66} & \underline{47.77} & \underline{63.59} & 68.42 & \underline{60.37} & 62.56 & \underline{61.33} & \underline{63.17} \\
         & \multirow{-2}{*}{Dense} & Time~$\downarrow$ & 15.2 & 28.2 & 18.4 & 27.4 & 7.5 & 15.1 & 7.2 & 95.4 & 26.8 & 4.1 & 3.6 & 6.2 & 15.0 & 7.2 \\
        \cline{2-17}
         &  & Acc~$\uparrow$ & 65.40 & 53.10 & 57.27 & 52.41 & 80.99 & 43.53 & 83.07 & 45.92 & 60.21 & 65.62 & 59.41 & 61.76 & 58.22 & 61.25 \\
         & \multirow{-2}{*}{FastV} & Time~$\downarrow$ & 14.3 & \underline{24.2} & \textbf{15.7} & \textbf{20.9} & \textbf{6.4} & \underline{15.0} & \textbf{6.1} & \underline{78.0} & \underline{22.6} & 3.4 & \textbf{2.5} & \textbf{4.6} & \textbf{10.5} & \textbf{5.2} \\
        \cline{2-17}
         &  & Acc~$\uparrow$ & 72.20 & 53.38 & 57.98 & 51.32 & \textbf{83.23} & 47.10 & 82.98 & 47.29 & 61.93 & \textbf{68.62} & 59.81 & 63.04 & 60.78 & 63.06 \\
         & \multirow{-2}{*}{VisionPulse} & Time~$\downarrow$ & 15.5 & 27.8 & 17.3 & 25.3 & 7.0 & 15.4 & 6.6 & 87.9 & 25.3 & \underline{3.2} & 3.3 & 5.6 & 13.9 & 6.5 \\
        \cline{2-17}
         &  & Acc~$\uparrow$ & 57.20 & 47.97 & 42.50 & 49.25 & 77.88 & 36.16 & 83.14 & 39.25 & 54.17 & 67.97 & \underline{60.37} & 61.44 & 55.22 & 61.25 \\
         & \multirow{-2}{*}{SparseVILA} & Time~$\downarrow$ & 15.4 & 25.9 & 17.5 & 23.7 & 7.0 & 16.8 & 6.6 & 83.4 & 24.5 & 3.3 & \underline{3.1} & \textbf{4.6} & 11.9 & 5.7 \\
        \cline{2-17}
         &  & Acc~$\uparrow$ & 61.10 & 50.84 & 47.11 & 53.05 & 78.01 & 41.52 & 83.50 & 44.49 & 57.45 & 65.28 & 58.15 & \textbf{65.28} & 59.00 & 61.93 \\
         & \multirow{-2}{*}{TransPrune} & Time~$\downarrow$ & \underline{13.7} & 28.5 & 17.5 & 22.5 & 7.2 & 21.3 & 6.5 & \textbf{68.8} & 23.2 & 3.6 & 3.3 & 5.3 & 12.4 & 6.2 \\
        \cline{2-17}
         &  & Acc~$\uparrow$ & 58.00 & 49.06 & 42.65 & 46.44 & 79.08 & 40.62 & 82.27 & 39.37 & 54.69 & 67.83 & 60.04 & \underline{64.16} & 55.67 & 61.92 \\
         & \multirow{-2}{*}{VisionZip} & Time~$\downarrow$ & 14.1 & 26.0 & 18.0 & 26.7 & 6.8 & 19.9 & 6.4 & 88.5 & 25.8 & \textbf{2.9} & \underline{3.1} & \textbf{4.6} & \underline{11.6} & \underline{5.6} \\
        \cline{2-17}
         &  & Acc~$\uparrow$ & 71.40 & 51.78 & 55.93 & 43.74 & 79.34 & 41.96 & 77.38 & 45.32 & 58.36 & 63.00 & 58.33 & 61.44 & 57.00 & 59.94 \\
         & \multirow{-2}{*}{QCache} & Time~$\downarrow$ & 15.3 & 29.6 & 19.1 & 24.3 & 8.7 & 16.4 & 7.3 & 131.8 & 31.6 & 6.6 & 8.1 & 12.1 & 20.8 & 11.9 \\
        \cline{2-17}
        \rowcolor{vaportint}
        \cellcolor{white} &  & Acc~$\uparrow$ & \textbf{73.60} & \textbf{55.43} & \textbf{63.11} & \textbf{54.89} & \underline{82.58} & \textbf{50.89} & \underline{83.64} & \textbf{48.48} & \textbf{64.08} & \underline{68.58} & \textbf{60.52} & 63.20 & \textbf{62.11} & \textbf{63.60} \\
        \rowcolor{vaportint}
        \cellcolor{white}\multirow{-16}{*}{\rotatebox{90}{\textbf{InternVL3.5-8B}}} & \multirow{-2}{*}{\textbf{VAPOR}} & Time~$\downarrow$ & \textbf{13.5} & \textbf{23.5} & \underline{15.9} & \underline{21.2} & \underline{6.5} & \textbf{12.6} & \underline{6.2} & 78.2 & \textbf{22.2} & 3.8 & 3.6 & \underline{5.1} & 13.3 & 6.5 \\
        \bottomrule
    \end{tabular}
    \endgroup
\end{table*}

\section{Experiments}
\label{sec:experiments}

\subsection{Experimental Setup}
\label{sec:setup}

\smallskip\noindent\textbf{Benchmarks and metrics.}
We evaluate VAPOR on eight image and four video reasoning benchmarks.
The image benchmarks cover visual mathematical reasoning (MathVista~\cite{lu2024mathvista}, MathVerse~\cite{zhang2024mathverse}, DynaMath~\cite{zou2025dynamath}, and WeMath~\cite{qiao2025we}), diagram understanding and visual logic (AI2D~\cite{kembhavi2016diagram} and LogicVista~\cite{xiao2024logicvista}), and general visual understanding and multidisciplinary reasoning (MMBench~\cite{liu2024mmbench} and MMMU-Pro~\cite{yue2025mmmu}).
For video reasoning, we use MVBench~\cite{li2024mvbench} for temporal understanding, Video-MME~\cite{fu2025video} for diverse video content, and MMVU~\cite{zhao2025mmvu} and Video-MMMU~\cite{hu2026video} for knowledge-intensive video reasoning.
We report accuracy and mean decoding latency per sample, with unweighted averages across benchmarks within each modality.
Additional benchmark details are provided in Sec.~\ref{app:benchmarks}.

\smallskip\noindent\textbf{Baselines.}
We compare VAPOR against standard full-attention decoding (\textbf{Dense}) and six representative acceleration methods.
These include visual token reduction methods (FastV~\cite{chen2024image}, TransPrune~\cite{li2026transprune}, and VisionZip~\cite{yang2025visionzip}), decoding-time visual selection methods (SparseVILA~\cite{khaki2025sparsevila} and VisionPulse~\cite{xu2026visionpulse}), and cross-layer attention reuse (QCache~\cite{zhuang2026q}).
Together, these baselines cover complementary strategies for reducing visual attention computation.
Their configurations are detailed in Sec.~\ref{app:baselines}.

\smallskip\noindent\textbf{Implementation details.}
We evaluate Qwen3-VL-8B-Instruct and InternVL3.5-8B with greedy decoding (8,192-token limit) and 64 uniformly sampled frames per video.
We set the proxy rank to $r=16$ for both models and the maximum cache age to $D=512$ for Qwen3-VL and $D=256$ for InternVL.
The input-level prefill consistency score $C$ is computed as the median across valid layer--head pairs.
For Qwen3-VL, the base visual-mass threshold is $\theta_M=0.2$ when $C<0.75$ and $0.1$ otherwise.
For InternVL, we use $\theta_M=0.15$ when $C<0.7$ and $0.05$ otherwise.
Most experiments are conducted on NVIDIA A100 40\,GB GPUs, with NVIDIA A800 80\,GB GPUs used for a small number of samples requiring additional memory.
For each model, all methods share the same sample-to-device assignments and evaluation settings.
Further implementation and measurement details are provided in Sec.~\ref{app:implementation}.

\subsection{Main Results}
\label{sec:main_results}

\Cref{tab:main_results} compares VAPOR with Dense decoding and six acceleration baselines across twelve image and video reasoning benchmarks.

\smallskip\noindent\textbf{Reasoning accuracy.}
VAPOR achieves the highest average accuracy across both modalities and model families, improving over Dense on nearly all individual benchmark comparisons. The gains are particularly notable on challenging reasoning tasks such as LogicVista, MMMU-Pro, and Video-MMMU. These results suggest that visual attention reuse can improve reasoning accuracy rather than merely preserve it under reduced computation.

\smallskip\noindent\textbf{Decoding efficiency.}
VAPOR reduces average decoding latency across all four model--modality settings, achieving 1.11--1.21$\times$ speedups over Dense. The improvements are particularly beneficial for computationally demanding image reasoning tasks. These results demonstrate that reusing visual attention outputs can reduce decoding cost without permanently discarding visual tokens.

\smallskip\noindent\textbf{Comparison with acceleration baselines.}
VAPOR achieves higher average accuracy than the strongest evaluated acceleration baseline in all four model--modality settings, with improvements of 0.35--2.15 percentage points. It also achieves lower or comparable decoding latency in every setting. This favorable accuracy--efficiency trade-off distinguishes VAPOR from existing approaches based primarily on visual token reduction or cross-layer attention reuse.

\subsection{Ablation Studies and Analysis}
\label{sec:ablations}

\subsubsection{Refresh Threshold and Cache Age}
\label{sec:ablation_threshold_age}
\begin{figure}[t]
    \centering
    \includegraphics[width=\columnwidth]{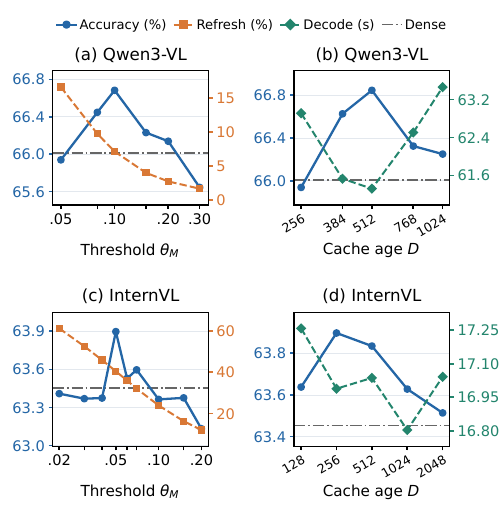}
    \caption{Sensitivity of accuracy, visual refresh rate, and decoding latency to the base visual-mass threshold $\theta_M$ and maximum cache age $D$. Gray dash-dotted lines indicate Dense accuracy.}
    \label{fig:threshold_age}
\end{figure}

\smallskip\noindent\textbf{Sensitivity to the refresh threshold and cache age.}
\Cref{fig:threshold_age} examines how the base visual-mass threshold $\theta_M$ and maximum cache age $D$ shape the balance between visual reuse and refresh.
We vary $\theta_M$ across all twelve benchmarks, fixing $D=512$ for Qwen3-VL and $D=256$ for InternVL. We then vary $D$, using the two-level threshold rule for Qwen3-VL and a fixed threshold of $0.05$ for InternVL.
As expected, increasing $\theta_M$ reduces the refresh rate because a larger threshold makes reuse more likely. Accuracy, however, does not improve monotonically with the refresh rate. It peaks at $\theta_M=0.10$ for Qwen3-VL and $\theta_M=0.05$ for InternVL.
The effect of cache age is also non-monotonic. The highest accuracy among the tested settings occurs at $D=512$ for Qwen3-VL and $D=256$ for InternVL. These results suggest that effective reuse depends not simply on refreshing more or less often, but on choosing an appropriate balance between persistence and timely updates.

\smallskip\noindent\textbf{Effect of the cache-age schedule.}
We next examine whether the shape of the age-dependent threshold decay affects performance. \Cref{tab:age_schedule} compares five schedules on Qwen3-VL with $D=512$, with their definitions and curves provided in Sec.~\ref{app:age_schedules}.
The cosine schedule gives the best result among the tested alternatives, reaching 66.85\% accuracy and 61.32 seconds per sample. Its refresh rate (4.75\%) is comparable to those of the other schedules. Accuracy nevertheless varies from 66.22\% to 66.85\%, suggesting that performance depends not only on how often Refresh occurs, but also on when it occurs.

\begin{table}[h]
\centering
\caption{Comparison of cache-age schedules on Qwen3-VL with $D=512$. Acc. denotes accuracy (\%), Refresh denotes refresh rate (\%), and Time denotes mean decoding latency (s/sample).}
\label{tab:age_schedule}
\small
\setlength{\tabcolsep}{4pt}
\begin{tabular}{lccc}
\toprule
Schedule & Acc. $\uparrow$ & Refresh $\downarrow$ & Time $\downarrow$ \\
\midrule
Constant & 66.22 & \textbf{4.56} & 63.36 \\
Linear & 66.28 & 4.84 & 62.54 \\
Raised cosine & 66.25 & 4.81 & 63.86 \\
Quadratic & 66.49 & 4.73 & 61.74 \\
\rowcolor{vaportint}
Cosine (ours) & \textbf{66.85} & 4.75 & \textbf{61.32} \\
\bottomrule
\end{tabular}
\end{table}

\begin{figure}[t]
    \centering
    \includegraphics[width=\columnwidth]{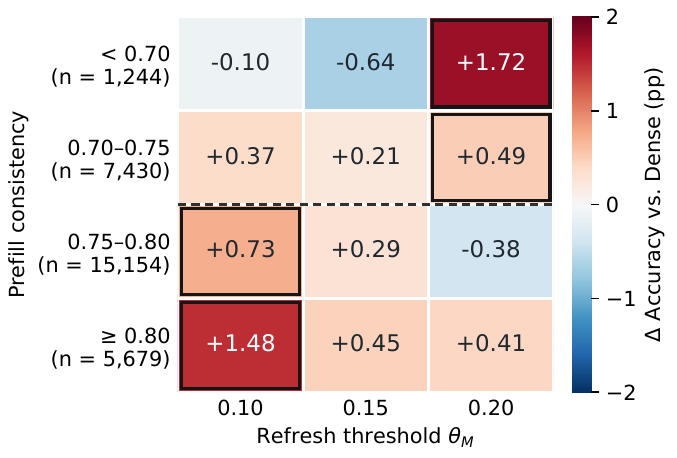}
    \caption{Accuracy change relative to Dense across prefill-consistency groups and base visual-mass thresholds on Qwen3-VL. The dashed line marks $C=0.75$.}
    \label{fig:consistency_threshold}
\end{figure}

\smallskip\noindent\textbf{Prefill consistency as a threshold-calibration signal.}
Prefill consistency provides an input-dependent cue for setting the base threshold.
\Cref{fig:consistency_threshold} shows the accuracy gains over Dense for different thresholds across prefill-consistency groups on Qwen3-VL.
Overall, the preferred threshold tends to decrease as prefill consistency increases, indicating that more consistent prompt-token attention favors a lower refresh threshold.
Based on this trend, we adopt a two-level calibration rule, setting $\theta_M=0.20$ for $C<0.75$ and $\theta_M=0.10$ otherwise.

\subsubsection{Accuracy of Proxy Visual-Mass Estimation}
\label{sec:ablation_rank}

We compare proxy and exact visual mass across six approximation ranks on 32 MathVista trajectories from Qwen3-VL-8B. For $N$ paired head--step observations, let $e_i=M_{P,i}-M_{E,i}$. We report Bias, the signed mean error, and MAE, the mean absolute error, both in percentage points.

\begin{figure}[t]
    \centering
    \includegraphics[width=\columnwidth]{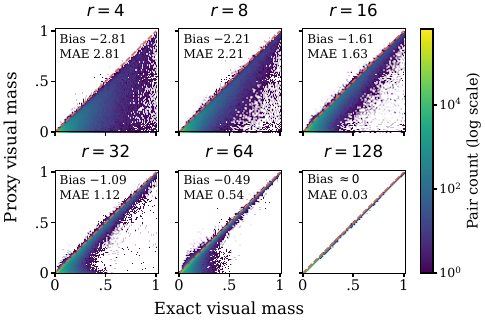}
    \caption{Proxy versus exact visual mass at matched head--step positions from 32 Qwen3-VL-8B MathVista trajectories.}
    \label{fig:rank_mass_density}
\end{figure}

As shown in \Cref{fig:rank_mass_density}, low-rank estimates systematically underestimate visual mass. Both the magnitude of the bias and MAE decrease as the rank increases, indicating improved estimation accuracy. At the default rank $r=16$, the bias and MAE are $-1.61$ and 1.63 percentage points, respectively.

This underestimation also affects threshold selection. Lower ranks tend to favor smaller thresholds to compensate for underestimated visual mass, although the relationship is not strictly monotonic (Sec.~\ref{app:rank_diagnostics}). Since rank and threshold jointly affect both reuse decisions and subsequent state transitions, more accurate mass estimation does not necessarily translate into higher reasoning accuracy.

\subsubsection{Visual Grounding during Reasoning}
\label{sec:contextual_interference}

\begin{figure}[t]
    \centering
    \includegraphics[width=\columnwidth]{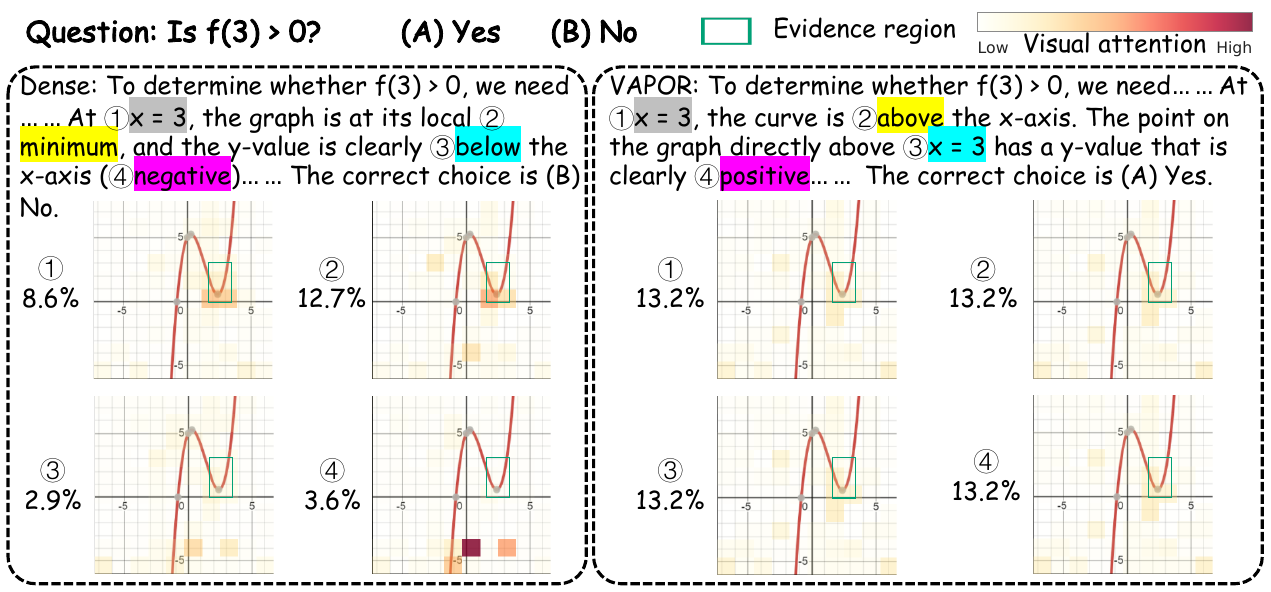}
    \caption{\textbf{Visual grounding during reasoning.}
    A MathVista example showing the reasoning traces and visual attention of Dense (left) and VAPOR (right).
    Numbered snapshots correspond to four reasoning positions at layer 12, head 12 of Qwen3-VL-8B.
    Green boxes mark the evidence region, and percentages indicate its share of visual attention.}
    \label{fig:visual_grounding}
\end{figure}

\Cref{fig:visual_grounding} illustrates how visual attention evolves during reasoning on a MathVista example using Qwen3-VL-8B.
The question asks whether the curve lies above the horizontal axis at $x=3$.
Dense gradually shifts attention away from the relevant evidence region, with its attention share decreasing from 8.6\% and 12.7\% at the first two positions to 2.9\% and 3.6\% at the last two, leading to an incorrect answer.
This suggests that token-level changes in the textual context can perturb visual routing, even when the reasoning does not require new visual evidence.
By recomputing visual attention at every step, Dense allows these changes to continuously influence which visual information receives attention.
In contrast, VAPOR maintains a consistent 13.2\% attention share by reusing its cached visual output and correctly identifies the positive value.
This reuse limits query-induced fluctuations in visual attention, helping preserve focus on task-relevant evidence as reasoning progresses.
Additional head-wise refresh statistics are provided in Sec.~\ref{app:head_refresh} of the supplementary material.

\begin{table}[t]
    \centering
    \caption{Prefill and decoding latency (s/sample) and generated output length with different methods for Qwen3-VL-8B. Best and second-best latency are shown in \textbf{bold} and \underline{underline}, respectively.}
    \label{tab:efficiency_breakdown}
    \begingroup
    \footnotesize
    \setlength{\tabcolsep}{0pt}
    \renewcommand{\arraystretch}{1.05}
    \resizebox{\columnwidth}{!}{%
    \begin{tabular}{@{}lrrr@{\hspace{5pt}}rrr@{}}
        \toprule
         & \multicolumn{3}{c}{\textbf{Image Reasoning}} & \multicolumn{3}{c}{\textbf{Video Reasoning}} \\
        \cmidrule(lr){2-4} \cmidrule(lr){5-7}
        Method & Prefill $\downarrow$ & Decode $\downarrow$ & Tokens $\downarrow$ & Prefill $\downarrow$ & Decode $\downarrow$ & Tokens $\downarrow$ \\
        \midrule
        Dense & 0.14 & 87.06 & 1710.6 & 1.25 & 37.09 & 728.5 \\
        FastV & 0.13 & \underline{79.50} & 1795.0 & 0.95 & 33.77 & 755.6 \\
        VisionPulse & 0.16 & 87.37 & 1714.5 & 1.31 & 36.25 & 709.7 \\
        SparseVILA & 0.21 & 99.68 & 2261.6 & 2.17 & 34.47 & 780.8 \\
        TransPrune & \underline{0.13} & 94.78 & 2122.1 & \textbf{0.63} & 34.53 & 771.3 \\
        VisionZip & \textbf{0.12} & 84.39 & 1913.6 & \underline{0.72} & \underline{32.95} & 739.5 \\
        QCache & 0.15 & 86.14 & 1709.0 & 1.27 & 34.98 & 693.4 \\
        \rowcolor{vaportint}
        \textbf{VAPOR} & 0.30 & \textbf{75.68} & \textbf{1694.9} & 1.50 & \textbf{32.59} & \textbf{684.0} \\
        \bottomrule
    \end{tabular}}
    \endgroup
\end{table}

\subsection{Efficiency Analysis}
\label{sec:efficiency_analysis}

Table~\ref{tab:efficiency_breakdown} reports prefill and decoding latency, along with generated output length, for Qwen3-VL. VAPOR generates fewer tokens than Dense on both image and video tasks, suggesting more concise and efficient reasoning.
Although low-rank preparation introduces modest prefill overhead, decoding dominates Dense runtime (99.8\% for image and 96.7\% for video), allowing VAPOR's decoding savings to outweigh this additional cost.
Overall, VAPOR achieves end-to-end speedups of 1.15$\times$ and 1.12$\times$ on image and video tasks, respectively.
Compared with other acceleration baselines, VAPOR achieves the lowest decoding latency in both modalities, outperforming the fastest competing baseline on image tasks (FastV) and video tasks (VisionZip).

\section{Conclusion}
\label{sec:conclusion}

In this paper, we presented \textbf{VAPOR}, a training-free framework for adaptive visual attention reuse during multimodal autoregressive decoding. VAPOR decouples visual attention updates from token generation through head-wise \emph{Persist} and \emph{Refresh} decisions, guided by low-rank visual mass estimation, prefill-based attention consistency, and cache-age control. By retaining the full visual KV cache, VAPOR reduces redundant visual attention computation without sacrificing access to visual information. Experiments demonstrate that VAPOR achieves a superior accuracy--efficiency trade-off over state-of-the-art acceleration baselines, while ablation studies validate the effectiveness of its key components and design choices. Our qualitative study further illustrates how maintaining visual attention across decoding steps can help preserve focus on task-relevant evidence. Overall, VAPOR establishes adaptive visual attention reuse as an effective strategy for achieving more accurate and efficient multimodal reasoning without sacrificing access to visual information.

\FloatBarrier
{
    \small
    \bibliographystyle{ieeenat_fullname}
    \bibliography{main}
}

\clearpage
\appendix

\section{Benchmark Details}
\label{app:benchmarks}

Our evaluation includes eight image and four video benchmarks.

\subsection{Image Reasoning}

\smallskip\noindent\textbf{MathVista.}
MathVista~\cite{lu2024mathvista} combines mathematical reasoning with visual interpretation across diverse tasks and image types, including diagrams, plots, and scientific figures.
Its questions require extracting relevant visual information and applying mathematical knowledge, rather than solving a text-only problem.
We evaluate all 1,000 questions in the \texttt{testmini} split.

\smallskip\noindent\textbf{MathVerse.}
MathVerse~\cite{zhang2024mathverse} evaluates mathematical reasoning grounded in visual diagrams.
The benchmark varies the amount of information conveyed through text and images to examine whether models actually use the diagram when solving a problem.
We evaluate 3,940 questions in the multimodal \texttt{testmini} split, covering the five image-containing variants.

\smallskip\noindent\textbf{DynaMath.}
DynaMath~\cite{zou2025dynamath} measures robustness to changes in visual mathematical problems, including numerical values and function graphs.
Its programmable seed questions generate related visual and textual variants, testing whether a model can apply the same reasoning to modified inputs.
We evaluate all ten released splits, \texttt{sample\_variant1} through \texttt{sample\_variant10}, totaling 5,010 questions.

\smallskip\noindent\textbf{WeMath.}
We-Math~\cite{qiao2025we} organizes visual mathematical problems by the knowledge concepts needed to solve them.
By relating composite problems to their component subproblems, it distinguishes knowledge acquisition from the ability to generalize and compose that knowledge in multi-step reasoning.
We evaluate all 1,740 questions in the \texttt{testmini} split and report answer accuracy.

\smallskip\noindent\textbf{AI2D.}
AI2D~\cite{kembhavi2016diagram} focuses on interpreting illustrative science diagrams and answering questions about their content.
Understanding diagram components, labels, and their relationships is necessary to recover the concepts or processes depicted.
We evaluate the standard 3,088-question test set, including letter-labeled diagrams used by the evaluation protocol.

\smallskip\noindent\textbf{LogicVista.}
LogicVista~\cite{xiao2024logicvista} tests logical reasoning in visual contexts, including navigation and puzzle solving.
It covers five logical reasoning tasks and nine capabilities, with questions requiring models to derive an answer from the visual information and task constraints.
We evaluate all 448 multiple-choice questions in the \texttt{test} split.

\smallskip\noindent\textbf{MMBench.}
MMBench~\cite{liu2024mmbench} provides a broad assessment of multimodal perception and reasoning through a structured taxonomy of abilities.
Its multiple-choice questions cover diverse visual understanding skills, complementing the more specialized mathematical and logical reasoning benchmarks.
We evaluate all 4,377 questions in the English development (\texttt{dev}) split.

\smallskip\noindent\textbf{MMMU-Pro.}
MMMU-Pro~\cite{yue2025mmmu} evaluates expert-level reasoning across academic disciplines with an emphasis on genuine visual understanding.
It strengthens the original MMMU evaluation by filtering questions answerable without images and increasing answer-choice difficulty.
It also introduces a vision-only setting.
We evaluate 1,679 questions from the \texttt{test} split under the standard 10-option setting.

\subsection{Video Reasoning}

\smallskip\noindent\textbf{MVBench.}
MVBench~\cite{li2024mvbench} comprises 20 tasks designed to require temporal understanding beyond a single frame.
Its multiple-choice questions test video perception and cognition, emphasizing information that must be integrated across time.
We evaluate the full released evaluation set of 4,000 questions across all 20 tasks.

\smallskip\noindent\textbf{Video-MME.}
Video-MME~\cite{fu2025video} covers diverse video content across six domains and 30 subfields, with short, medium, and long videos ranging from seconds to an hour.
Its questions assess understanding across different scenes and temporal scales.
We evaluate all 2,700 questions in the \texttt{test} split using visual frames, without audio or subtitles.

\smallskip\noindent\textbf{MMVU.}
MMVU~\cite{zhao2025mmvu} evaluates expert-level video understanding across 27 subjects in science, healthcare, humanities and social sciences, and engineering.
The questions require combining observations from the video with domain knowledge, extending evaluation beyond basic event recognition.
We evaluate all 625 multiple-choice questions in the \texttt{validation} split.

\smallskip\noindent\textbf{Video-MMMU.}
Video-MMMU~\cite{hu2026video} measures knowledge acquisition from professional videos across six disciplines.
Its questions target perception, comprehension, and adaptation, including applying knowledge learned from a video to a new problem.
We evaluate the full released evaluation set of 900 questions across the perception, comprehension, and adaptation tracks, reporting answer accuracy rather than the benchmark's separate knowledge-gain metric.

\section{Baseline Details}
\label{app:baselines}

For baselines requiring hyperparameter selection, we evaluate candidate configurations on the full evaluation samples and prioritize accuracy when selecting the final configuration. Method-specific tie-breaking criteria, when applicable, are described alongside the corresponding baselines. The selected configuration is fixed across benchmarks for each model, rather than tuned separately for individual benchmarks. Baselines without additional hyperparameter searches follow their reference configurations.

\smallskip\noindent\textbf{Dense.}
Dense retains all visual tokens and recomputes attention over the complete available visual and textual context at every decoding step.
It provides the unmodified full-attention reference for each model.

\subsection{Visual Token Compression}

\smallskip\noindent\textbf{FastV.}
FastV~\cite{chen2024image} uses attention in early language-model layers to identify redundant visual tokens and prunes them for subsequent layers.
On Qwen3-VL, we first compare $K\in\{2,4,8\}$ at a keep ratio of 0.50 and select $K=4$.
We then compare keep ratios $\{0.10,0.25,0.50,0.75\}$ and select 0.75 by accuracy.
InternVL uses $K=4$ and a fixed keep ratio of 0.25.

\smallskip\noindent\textbf{TransPrune.}
TransPrune~\cite{li2026transprune} progressively prunes visual tokens using changes in their representations across layers, together with instruction-guided attention.
Its transition criterion considers both the magnitude and direction of representation changes, providing a token-importance signal beyond attention alone.
We adopt the reference retention schedule $(0.875,0.625,0.125)$ at stage indices $(7,9,12)$ for both models.

\smallskip\noindent\textbf{VisionZip.}
VisionZip~\cite{yang2025visionzip} compresses visual representations before they enter the language model.
It retains dominant tokens identified from vision-encoder attention and merges other tokens by semantic similarity into contextual tokens, preserving complementary information in a shorter visual sequence.
We adopt the reference implementation's 50\% retention setting, with dominant and contextual ratios of 0.45 and 0.05, respectively, for both models.

\subsection{Decoding-time Visual Selection}

\smallskip\noindent\textbf{SparseVILA.}
SparseVILA~\cite{khaki2025sparsevila} decouples visual sparsity between prefill and decoding. It prunes redundant tokens during prefill and retrieves query-relevant tokens during decoding, retaining visual cache entries for subsequent retrieval. For both Qwen3-VL and InternVL3.5, we evaluate five prefill/decoding keep-ratio pairs: \((0.25, 0.10)\), \((0.30, 0.15)\), \((0.40, 0.25)\), \((0.50, 0.10)\), and \((0.50, 0.25)\). We prioritize accuracy and use total inference time to break ties. The selected configuration for both models is \((0.50, 0.25)\).

\smallskip\noindent\textbf{VisionPulse.}
VisionPulse~\cite{xu2026visionpulse} selects visual tokens separately at each reasoning step as the required evidence changes.
It uses a lightweight visual-attention-mass signal to determine the retention budget, then retains the most relevant tokens under that budget.
For Qwen3-VL, we calibrate the anchor layer among indices $\{8,17,26\}$ and select 26.
With this layer fixed and method temperature 0.4, we compare budget ranges 0.05--0.10, 0.10--0.20, 0.20--0.40, and 0.40--0.80 and select 0.05--0.10 by accuracy.
InternVL uses anchor layer 17 with the same selected budget range and temperature.

\subsection{Cross-layer Attention Reuse}

\smallskip\noindent\textbf{QCache.}
QCache~\cite{zhuang2026q} exploits similar attention patterns across adjacent decoding layers. Its Lazy Attention mechanism shares queries through a layer-shared cache, reducing redundant attention computation across layers. For both models, we evaluate divergence thresholds \(\epsilon\in\{0.01,0.02,0.05,0.10\}\) and select \(\epsilon=0.01\), which achieves the highest accuracy.

\section{Cache-age Schedules}
\label{app:age_schedules}

\begin{figure}[t]
    \centering
    \includegraphics[width=\columnwidth]{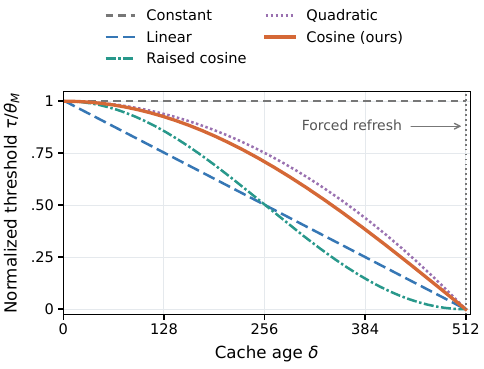}
    \caption{Cache-age schedules used in Table~\ref{tab:age_schedule}. The horizontal axis is cache age $\delta$ and the vertical axis is the threshold normalized by $\theta_M$. The dotted vertical line marks the mandatory refresh boundary at $D=512$.}
    \label{fig:age_schedules}
\end{figure}

The schedules compared in Table~\ref{tab:age_schedule} scale the base threshold as $\tau_{h,t}=\theta_M g(s)$, where $s=\operatorname{clip}((\delta_{h,t}-2)/(D-2),0,1)$ follows Eq.~\ref{eq:age_threshold}.
The five scaling functions are
\begin{equation}
\begin{aligned}
    g_{\mathrm{constant}}(s) &= 1, \\
    g_{\mathrm{linear}}(s) &= 1-s, \\
    g_{\mathrm{raised\ cosine}}(s) &= \tfrac{1}{2}(1+\cos(\pi s)), \\
    g_{\mathrm{quadratic}}(s) &= 1-s^2, \\
    g_{\mathrm{cosine}}(s) &= \cos(\tfrac{\pi}{2}s).
\end{aligned}
\end{equation}
Fig.~\ref{fig:age_schedules} visualizes these functions for $D=512$.
All schedules retain the same mandatory Refresh at $\delta_{h,t}\geq D$, including the constant schedule.
Only the age-dependent scaling function changes in this comparison.

\section{Head-wise Refresh Behavior}
\label{app:head_refresh}

We analyze all 1,152 query heads of Qwen3-VL-8B-Instruct on 96 trajectories, with 32 randomly selected examples each from MathVista, LogicVista, and MMVU.
We record Refresh decisions and exact visual mass every eight decoding steps through step 512, or until generation ends.
For each head, refresh frequency is the fraction of these sampled positions executed in Refresh, averaged equally over examples.
The mandatory initialization steps are excluded by this sampling schedule.
Exact mass is measured using the current query and keys along the VAPOR trajectory, so the comparison uses the same attention state rather than a separate Dense rollout.

\begin{figure*}[t]
    \centering
    \includegraphics[width=\textwidth]{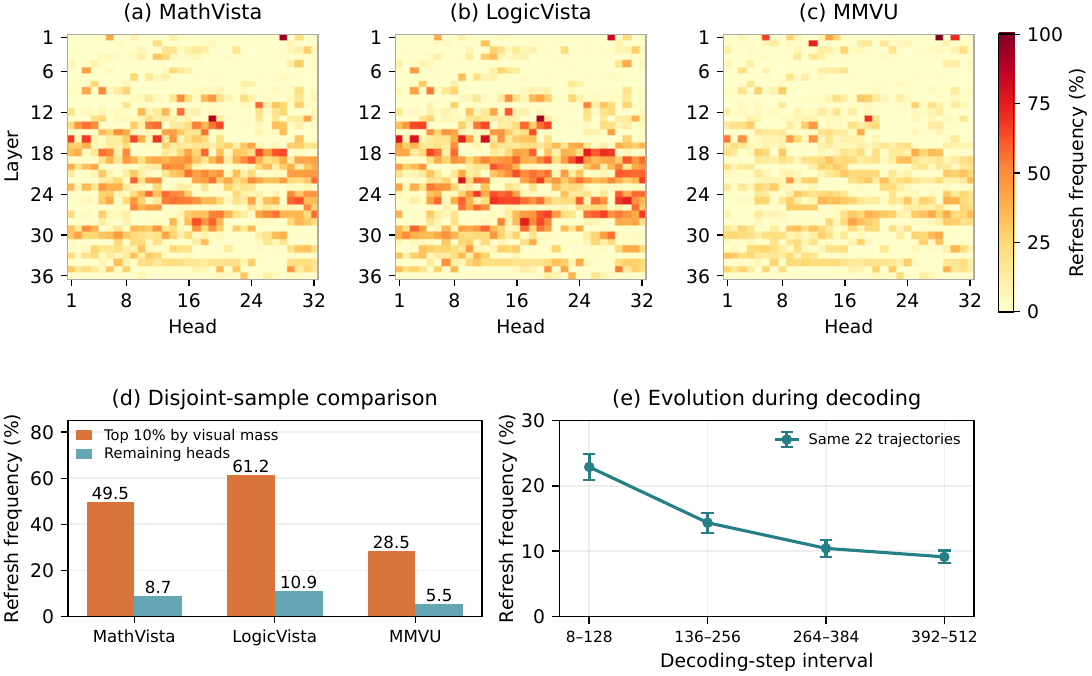}
    \caption{Head-wise refresh behavior of Qwen3-VL-8B-Instruct.
    (a--c) Refresh frequency by layer and query head, with one-based indices and a shared color scale.
    (d) Refresh frequency on 16 evaluation examples per benchmark for the top 10\% of heads ranked by mean exact visual mass on 16 disjoint examples, compared with the remaining heads.
    (e) Mean refresh frequency over four decoding-step intervals for the same 22 trajectories reaching step 512.
    Error bars indicate the standard error across trajectories.}
    \label{fig:head_refresh}
\end{figure*}

\smallskip\noindent\textbf{Frequency across heads.}
Fig.~\ref{fig:head_refresh}(a--c) shows substantial variation both within and across layers.
Mean refresh frequencies are 12.3\%, 15.2\%, and 7.9\% on MathVista, LogicVista, and MMVU, respectively, whereas the corresponding medians across heads are only 3.0\%, 4.1\%, and 1.9\%.
For each benchmark, we rank heads by refresh frequency averaged equally over all 32 examples and select the top 116 heads (approximately 10\%).
Their activity share is the sum of these mean frequencies over the selected heads divided by the sum over all heads, yielding 41.0\%, 38.8\%, and 41.5\% on MathVista, LogicVista, and MMVU, respectively.
This statistic uses refresh-based ranking on all 32 examples, whereas Fig.~\ref{fig:head_refresh}(d) uses mass-based ranking on 16 examples and evaluates on the other 16.
Layer-averaged frequency peaks at layer 19 for both image benchmarks and layer 25 for MMVU.
Thus, VAPOR concentrates visual updates in a subset of heads while retaining task-dependent differences, rather than assigning a uniform refresh budget.

\smallskip\noindent\textbf{Relation to visual attention.}
Across heads, mean exact visual mass correlates strongly with refresh frequency, with Spearman correlations of 0.914, 0.924, and 0.758 on the three benchmarks.
To assess whether the association transfers across examples, we identify the top 10\% of heads by visual mass using half of each benchmark's examples and measure their refresh frequency on the other half.
These heads refresh 5.2--5.7 times as often as the remaining heads (Fig.~\ref{fig:head_refresh}(d)).
This is consistent with prior findings that visual attention is unevenly distributed across heads and layers~\cite{bi2025unveiling}.
Our result shows that mass-based routing retains this heterogeneity in its allocation of computation over time.
However, high visual mass alone does not identify the object-localizing heads studied in visual grounding~\cite{kang2025your}.

The agreement is therefore qualitative, rather than a match to specific head identities across model families.

\smallskip\noindent\textbf{Evolution during decoding.}
We use the same 22 trajectories reaching step 512 in all four intervals to avoid changes in sample composition as shorter generations end.
This subset contains 3 MathVista, 16 LogicVista, and 3 MMVU examples.
Its mean refresh frequency decreases from 22.9\% in the first interval to 9.1\% in the last (Fig.~\ref{fig:head_refresh}(e)).
The pooled decline is not universal: MMVU increases after the second interval, although its final rate remains below its initial rate.
VAPOR therefore allows visual updates to become less frequent as decoding proceeds without imposing a decreasing schedule over absolute decoding position.
The threshold schedule instead depends on cache age and resets after each Refresh.
These observations describe adaptive refresh behavior and do not establish a causal effect of positional encoding.

\smallskip\noindent\textbf{Refresh decision granularity.}
We compare four policies on all 448 LogicVista test questions using Qwen3-VL-8B-Instruct (Table~\ref{tab:refresh_policy}).
All use $r=16$, $D=512$, and the original consistency-based threshold rule: $\theta_M=0.20$ when $C<0.75$ and $0.10$ otherwise.
Thresholds are identical across policies for each input.

\emph{Global} shares one refresh decision across all layers and heads, based on their mean visual mass at the preceding decoding step.
\emph{Layer-wise} instead averages the preceding-step masses within each layer and shares a decision among that layer's heads.
\emph{Head-wise (delayed refresh)} uses each head's own preceding-step mass independently.
These three controls use exact mass on Refresh steps and proxy mass on Persist steps, with a common one-step decision delay.
\emph{VAPOR} also makes independent head-wise decisions, but checks the current proxy mass during Persist and can refresh immediately at that same step.
The delayed head-wise control isolates the effect of decision timing while retaining independent head-wise routing.

\begin{table}[t]
    \centering
    \small
    \setlength{\tabcolsep}{3pt}
    \renewcommand{\arraystretch}{1.12}
    \caption{Refresh policies on all 448 LogicVista test questions.
    Refresh rate is the fraction of decoding layer--head positions that execute exact visual attention, pooled over complete generations and including initialization.
    Decode time is averaged over questions.}
    \label{tab:refresh_policy}
    \begin{tabular*}{\columnwidth}{@{\extracolsep{\fill}}lrrr@{}}
        \toprule
        Policy & Acc. (\%) $\uparrow$ & Refresh (\%) & Decode (s) $\downarrow$ \\
        \midrule
        Global & 28.35 & 1.07 & 248.97 \\
        Layer-wise & 46.65 & 2.59 & 193.21 \\
        \shortstack[l]{Head-wise\\(delayed refresh)} & 51.56 & 3.54 & 153.99 \\
        \midrule
        VAPOR & \textbf{52.46} & 4.32 & \textbf{135.91} \\
        \bottomrule
    \end{tabular*}
\end{table}

Under the shared-threshold protocol and a common one-step delay, the Head-wise (delayed refresh) control improves accuracy by 23.21 and 4.91 percentage points over Global and Layer-wise, respectively, supporting the value of head-specific visual updates.
VAPOR further reduces decoding latency by 11.7\% relative to the delayed head-wise control while maintaining comparable accuracy (52.46\% versus 51.56\%).

Table~\ref{tab:refresh_policy} counts actual exact-attention executions, including immediate proxy-triggered refreshes, rather than the Refresh state assigned to the next step.
In contrast, Fig.~\ref{fig:head_refresh} averages execution frequencies equally over examples using every eighth step within the first 512 decoding steps, excluding initialization.
The different horizons, example sets, and aggregation rules prevent direct comparison of their reported frequencies.

\section{Additional Implementation Details}
\label{app:implementation}

\smallskip\noindent\textbf{Runtime configuration.}
We implement evaluation with PyTorch 2.8.0 (CUDA 12.8), Hugging Face Transformers 4.57.6, and Triton 3.4.0 under Python 3.12.8.
Both models run in evaluation mode under \texttt{torch.inference\_mode}, with BF16 parameters and batch size one.
Generation uses dynamic KV caching and stops at EOS or the 8,192-token limit.

\smallskip
\noindent\textbf{Attention backends.}
The vision encoders use PyTorch scaled dot-product attention (SDPA).
For language-model attention, VAPOR follows the same attention backend
and numerical configuration as Dense, ensuring a consistent
implementation setting across the two methods. The difference lies
in the decoding strategy: Dense recomputes visual attention at every
step, whereas VAPOR adaptively reuses cached visual attention outputs
between refreshes while keeping textual attention up to date.
For memory-bounded eager prefill, the query block size is 512,
retaining all available keys and values and the original attention
mask. Qwen3-VL also bounds long-input FP32 softmax temporaries to
128 query rows when the query length exceeds 8,192.

\smallskip\noindent\textbf{Input preprocessing.}
Qwen3-VL uses its native processor, with video frames resized under a maximum per-frame pixel budget of 131,072.
InternVL uses $448\times448$ image tiles, with up to 12 dynamic tiles and an additional thumbnail when multiple tiles are used.
InternVL represents each video frame with a single $448\times448$ tile.
Within each model, prompt text, source frames, and preprocessing settings are fixed across methods.
Video inputs contain the same 64 uniformly sampled frames without audio or subtitles.

\smallskip\noindent\textbf{Latency measurement.}
Decoding latency starts after the first model forward and ends when generation and pending method-specific computation have completed.
CUDA synchronization delimits the measured interval.
We exclude model loading, warm-up, CPU input preparation, answer scoring, and output serialization.
Each method generates its own output up to EOS or the common token limit.
Accuracy and latency are averaged equally over benchmarks within each modality.

\subsection{Evaluation Prompts}
\label{app:eval_prompts}

We use the same benchmark-specific task instructions for Dense, VAPOR, and all acceleration baselines.
The templates below show the user-message text before applying each model's native chat template.
Square-bracketed fields denote sample-specific content, and the option lists extend to all supplied choices in their original order.
Angle-bracketed answer fields are literal instructions included in the prompt.
Dataset-provided question text and hints are retained rather than replaced by a generic question-only prompt.

\newcommand{\vaporevalprompt}[1]{%
    \par\smallskip\noindent
    \fbox{\begin{minipage}{\dimexpr\linewidth-2\fboxsep-2\fboxrule\relax}
        \small\ttfamily\raggedright #1
    \end{minipage}}\par\smallskip
}

\smallskip\noindent\textbf{MathVista.}
We retain the dataset's supplied \texttt{query}, including its answer-format hints and choices when present, and use the following template after the input image:
\vaporevalprompt{%
You are a careful visual mathematical reasoning model.\par
Reason through the visual evidence step by step.\par\medskip
[dataset query]\par\medskip
End with exactly:\par
FINAL: <answer>
}

\smallskip\noindent\textbf{MathVerse, WeMath, DynaMath, LogicVista, and MMBench.}
These benchmarks share the following wrapper after the input image:
\vaporevalprompt{%
[benchmark question text]\par\medskip
Reason carefully and end with exactly:\par
FINAL: <answer>
}
For MathVerse, the question text is the supplied \texttt{query\_cot} field.
For WeMath, it is the question followed by \texttt{Choices: [supplied option string]} on a new line.
DynaMath and LogicVista retain their supplied question text.
For MMBench, we concatenate the optional hint, question, and available choices, separating these parts with blank lines and formatting each choice as \texttt{(A) [option text]} on its own line.

\smallskip\noindent\textbf{AI2D.}
The task text lists the choices and requests an option letter, followed by the same shared image wrapper:
\vaporevalprompt{%
Question: [question]\par
Options:\par
A. [option A]\par
B. [option B]\par
[remaining options]\par
Solve the problem and give your final answer as FINAL: <option letter>.\par\medskip
Reason carefully and end with exactly:\par
FINAL: <answer>
}

\smallskip\noindent\textbf{MMMU-Pro.}
Images are inserted at their referenced positions within the question, preserving the ordering of multiple images.
The text template is
\vaporevalprompt{%
You are a careful visual reasoning model.\par
Reason through the visual evidence step by step and rely on the image whenever needed.\par\medskip
Question: [question with image positions]\par\medskip
Choices:\par
A. [option A]\par
B. [option B]\par
[remaining options]\par\medskip
Answer with the option letter.\par\medskip
End with exactly:\par
FINAL: <answer>
}
For vision-only questions, the question field consists of the supplied image followed by \texttt{Answer the question shown in the image.}

\smallskip\noindent\textbf{Video multiple-choice questions.}
MVBench, Video-MME, MMVU, and multiple-choice Video-MMMU questions use the following text after the visual input:
\vaporevalprompt{%
Watch the video and answer the following multiple-choice question.\par
[question]\par
Options:\par
(A) [option A]\par
(B) [option B]\par
[remaining options]\par
Think step by step, then end with a line in the format Answer: X, where X is the option letter.
}

\smallskip\noindent\textbf{Video open-ended questions.}
Open-ended Video-MMMU questions use
\vaporevalprompt{%
Watch the video and answer the following question.\par
[question]\par
Think step by step, then end with Answer: <answer>.
}

\smallskip\noindent\textbf{Model-specific visual and chat formatting.}
Qwen3-VL receives images or the 64 sampled video frames through its native multimodal processor, which renders the user message and assistant-generation prefix.
InternVL uses its native conversation template and default system message.
For videos, its user message begins with \texttt{Frame1: <image>}, through \texttt{Frame64: <image>}, each on a separate line, followed by the task text above.
When a video question supplies an additional image, that image is included after the video frames and before the question text.
Numbered image references in video question text are removed after retaining the supplied image as visual input.
These model-specific wrappers are fixed across all compared methods.

\subsection{Computational and Memory Overhead}
\label{app:memory_overhead}

\smallskip\noindent\textbf{Measurement protocol.}
We randomly sample 88 LogicVista test inputs and compare Dense and VAPOR with Qwen3-VL-8B-Instruct on A100 40GB GPUs.
Each pair uses the same GPU and input, with randomized method order and a fixed length of 512 generated tokens to control for output-length differences.
VAPOR uses the default $r=16$, $D=512$, and consistency-guided threshold rule.
After warm-up, we measure peak PyTorch-allocated GPU memory, including model weights, separately for prefill and decoding.
The prefill boundary is completion of the first model forward, and the request peak is the larger of the two stage peaks.
Allocator-reserved but unused memory is excluded.

\begin{table}[t]
    \centering
    \small
    \setlength{\tabcolsep}{4pt}
    \renewcommand{\arraystretch}{1.12}
    \caption{Mean per-request peak allocated memory (GiB) on 88 LogicVista inputs with 512 generated tokens.
    All columns include model weights.}
    \label{tab:memory_overhead}
    \begin{tabular*}{\columnwidth}{@{\extracolsep{\fill}}lrrr@{}}
        \toprule
        Method & Overall peak & Prefill peak & Decode peak \\
        \midrule
        Dense & 16.618 & 16.597 & 16.556 \\
        VAPOR & 16.637 & 16.598 & 16.595 \\
        \bottomrule
    \end{tabular*}
\end{table}

\smallskip\noindent\textbf{Storage overhead.}
VAPOR retains the full visual KV cache.
Its low-rank factors are shared by query heads belonging to the same KV head, while cached conditional visual outputs remain specific to each query head.
For $L$ layers, $H_{\mathrm{KV}}$ KV heads, and $H$ query heads, their FP32 storage is
\begin{equation*}
    \underbrace{4LH_{\mathrm{KV}}r(N_V+d_h)}_{\text{low-rank factors}}
    +\underbrace{4LHd_h}_{\text{cached visual outputs}}
    \quad\text{bytes}.
\end{equation*}
With $L=36$, $H_{\mathrm{KV}}=8$, $H=32$, and $d_h=128$, the measured factors occupy 8.97 MiB on average (3.39--32.48 MiB), and cached outputs occupy 0.56 MiB.
No $N_V$-dimensional attention distribution is stored per query head.
The implementation additionally retains key snapshots for asynchronous preparation and allocates routing and scratch buffers, all included in Table~\ref{tab:memory_overhead}.
Overall peak memory increases by 19.3 MiB on average, and decoding peak memory increases by 40.7 MiB.
The corresponding mean paired increases are 0.11\% and 0.24\%, respectively.

\begin{table}[t]
    \centering
    \small
    \setlength{\tabcolsep}{3pt}
    \renewcommand{\arraystretch}{1.12}
    \caption{Storage versus visual-token count on the same inputs, sorted by $N_V$ into four equal groups of 22.
    Values are group means. $\Delta$ Decode is VAPOR's additional decoding peak relative to Dense, including all runtime allocations.}
    \label{tab:memory_visual_length}
    \begin{tabular*}{\columnwidth}{@{\extracolsep{\fill}}rrrr@{}}
        \toprule
        Mean $N_V$ & \shortstack{Factors\\(MiB)} & \shortstack{Cached outputs\\(MiB)} & \shortstack{$\Delta$ Decode\\(MiB)} \\
        \midrule
        124 & 4.42 & 0.56 & 20.91 \\
        256 & 6.75 & 0.56 & 30.69 \\
        392 & 9.13 & 0.56 & 40.72 \\
        757 & 15.56 & 0.56 & 70.55 \\
        \bottomrule
    \end{tabular*}
\end{table}

The sampled inputs contain 65--1,720 visual tokens.
Table~\ref{tab:memory_visual_length} shows the linear growth of factor storage with $N_V$, whereas output-cache storage is independent of $N_V$.
These measurements characterize image inputs in this range, rather than long-video memory requirements.

\smallskip\noindent\textbf{Preparation and computation.}
We construct the truncated factors in FP32 through eigendecomposition of $K_V^\top K_V$ and projection onto its leading $r$ eigenvectors.
Preparation uses eight CUDA streams and overlaps the initial decoding steps.
Its mean elapsed span is 0.244 s, measured between the earliest start and latest completion events across layers.
This span is already included in decoding time and is not additive latency.
Mean prefill time is 0.122 s for Dense and 0.267 s for VAPOR, including its consistency calibration.
On the same fixed-length inputs, mean decoding time is 26.14 s for Dense and 22.95 s for VAPOR, including preparation and routing.

\begin{table*}[!t]
\centering
\caption{Reasoning accuracy and proxy approximation quality across
24 rank--threshold configurations. Accuracy is evaluated on the same
512 randomly sampled examples from the MathVista testmini split for
all configurations. Mass MAE and decision agreement are evaluated on
16 examples randomly selected from this subset, with each configuration
following its own decoding trajectory. Mass MAE is reported in
percentage points, and accuracy and decision agreement are reported
as percentages. Bold indicates the best result within each proxy
rank for each metric.}
\label{tab:rank_threshold_grid}
\small

\begin{minipage}[t]{0.325\textwidth}
\centering
\textbf{(a) Accuracy (\%) $\uparrow$}
\par\smallskip
\setlength{\tabcolsep}{1.6pt}
\begin{tabular*}{\linewidth}{@{\extracolsep{\fill}}rcccccc@{}}
\toprule
$\theta_M$ & $4$ & $8$ & $16$ & $32$ & $64$ & $128$ \\
\midrule
0.05 & \textbf{75.00} & \textbf{77.15} & 75.78 & 75.00 & 75.78 & 75.59 \\
0.10 & 73.63 & 75.59 & \textbf{77.54} & \textbf{75.78} & 75.20 & 75.78 \\
0.15 & 74.61 & 75.20 & 74.61 & 75.39 & \textbf{76.17} & \textbf{76.17} \\
0.20 & 73.83 & 73.44 & 75.00 & 75.39 & 74.80 & 75.00 \\
\bottomrule
\end{tabular*}
\end{minipage}
\hfill
\begin{minipage}[t]{0.325\textwidth}
\centering
\textbf{(b) Mass MAE (pp) $\downarrow$}
\par\smallskip
\setlength{\tabcolsep}{1.6pt}
\begin{tabular*}{\linewidth}{@{\extracolsep{\fill}}rcccccc@{}}
\toprule
$\theta_M$ & $4$ & $8$ & $16$ & $32$ & $64$ & $128$ \\
\midrule
0.05 & \textbf{1.860} & \textbf{1.499} & \textbf{1.107} & \textbf{0.610} & \textbf{0.206} & \textbf{0.010} \\
0.10 & 2.430 & 2.063 & 1.514 & 0.873 & 0.343 & 0.020 \\
0.15 & 2.544 & 2.125 & 1.618 & 0.964 & 0.388 & 0.024 \\
0.20 & 2.772 & 2.194 & 1.664 & 1.001 & 0.390 & 0.028 \\
\bottomrule
\end{tabular*}
\end{minipage}
\hfill
\begin{minipage}[t]{0.325\textwidth}
\centering
\textbf{(c) Decision agreement (\%) $\uparrow$}
\par\smallskip
\setlength{\tabcolsep}{1.6pt}
\begin{tabular*}{\linewidth}{@{\extracolsep{\fill}}rcccccc@{}}
\toprule
$\theta_M$ & $4$ & $8$ & $16$ & $32$ & $64$ & $128$ \\
\midrule
0.05 & 82.75 & 85.48 & 88.39 & 92.92 & 97.29 & 99.89 \\
0.10 & 89.18 & 89.89 & 92.80 & 95.60 & 98.17 & 99.88 \\
0.15 & 93.23 & 94.34 & 96.03 & 97.66 & 99.13 & 99.94 \\
0.20 & \textbf{94.85} & \textbf{96.47} & \textbf{97.61} & \textbf{98.72} & \textbf{99.54} & \textbf{99.96} \\
\bottomrule
\end{tabular*}
\end{minipage}

\end{table*}

\section{Visual Mass Approximation Bound}
\label{app:mass_bound}

We prove Proposition~\ref{prop:mass_bound} for a fixed attention head and decoding step.
Queries and keys include positional encoding, as in Sec.~\ref{sec:attention_output}.
The exact and approximate computations use the same valid keys, attention mask, and text partition sum $Z_{T,t}>0$.
All quantities in the proof are evaluated in exact arithmetic.

\smallskip\noindent\textbf{From key approximation to logit error.}
Let $R_r=K_V-\widehat K_V$ and define
\begin{equation}
    \varepsilon_t
    =\|\hat z_{V,t}-z_{V,t}\|_\infty
    =\frac{\|R_rq_t\|_\infty}{\sqrt{d_h}}.
    \label{eq:app_logit_error}
\end{equation}
Any additive attention mask is identical in both logits and cancels in their difference on valid positions.
Writing the discarded singular components as $R_r=\sum_{j>r}\sigma_j u_jv_j^{\mathsf T}$ gives
\begin{equation}
\begin{aligned}
    \|R_rq_t\|_2^2
    &=\sum_{j>r}\sigma_j^2(v_j^{\mathsf T}q_t)^2\\
    &\leq\sigma_{r+1}^2\|q_t\|_2^2.
\end{aligned}
\end{equation}
Thus, using $\|\cdot\|_\infty\leq\|\cdot\|_2$,
\begin{equation}
    \varepsilon_t\leq
    \frac{\sigma_{r+1}(K_V)\|q_t\|_2}{\sqrt{d_h}}.
    \label{eq:app_spectral_error}
\end{equation}
We take $\sigma_{r+1}=0$ when the retained components span the full rank.

\smallskip\noindent\textbf{From logit error to mass error.}
Let $\widehat Z_{V,t}=\sum_{i\in\mathcal V}\exp(\hat z_{V,t,i})$.
The elementwise logit bound implies
\begin{equation}
    e^{-\varepsilon_t}Z_{V,t}
    \leq\widehat Z_{V,t}
    \leq e^{\varepsilon_t}Z_{V,t}.
\end{equation}
Therefore, for $\Delta_t=\log\widehat Z_{V,t}-\log Z_{V,t}$, we have $|\Delta_t|\leq\varepsilon_t$.
Define $x_t=\log Z_{V,t}-\log Z_{T,t}$ and $s(x)=(1+e^{-x})^{-1}$.
The two masses satisfy
\begin{equation}
    M_{E,t}=s(x_t),\qquad M_{P,t}=s(x_t+\Delta_t).
\end{equation}
For a nonnegative shift $a$, $s(x+a)-s(x)$ is maximized at $x=-a/2$, where it equals $\tanh(a/4)$.
The same bound holds for a negative shift by exchanging the two arguments.
Consequently,
\begin{equation}
\begin{aligned}
    |M_{P,t}-M_{E,t}|
    &\leq\tanh\!\left(\frac{|\Delta_t|}{4}\right)\\
    &\leq\tanh\!\left(\frac{\varepsilon_t}{4}\right)
    \leq\frac{\varepsilon_t}{4}
    \leq b_{r,t}.
\end{aligned}
    \label{eq:app_tight_mass_bound}
\end{equation}
This proves Proposition~\ref{prop:mass_bound}.

\smallskip\noindent\textbf{Threshold-decision stability.}
Let $\tau_{h,t}$ be the current threshold, including any cache-age adjustment.
If $M_{E,t}>\tau_{h,t}+b_{r,t}$, then $M_{P,t}>\tau_{h,t}$.
If $M_{E,t}<\tau_{h,t}-b_{r,t}$, then $M_{P,t}<\tau_{h,t}$.
Hence
\begin{equation}
\begin{aligned}
    |M_{E,t}-\tau_{h,t}|>b_{r,t}\quad\Longrightarrow\quad\\
    \mathbf{1}[M_{P,t}\geq\tau_{h,t}]
    =\mathbf{1}[M_{E,t}\geq\tau_{h,t}].
\end{aligned}
\end{equation}
The same implication holds with the tighter margin $\tanh(\varepsilon_t/4)\leq b_{r,t}$.
This comparison holds at the same query and cache state, with other refresh conditions fixed.
It does not assert that separate decoding runs follow identical trajectories.

\smallskip\noindent\textbf{Scope of the guarantee.}
The bound connects discarded singular values to visual-mass error.
Truncated SVD minimizes the spectral-norm matrix residual, but need not minimize the nonlinear mass error for a particular query.
A larger rank reduces the spectral upper bound for fixed keys and query, without guaranteeing higher downstream accuracy.
For a numerically computed factorization, $\sigma_{r+1}(K_V)$ can be replaced by the actual spectral residual.
If finite-precision mass evaluations introduce absolute errors bounded by $\eta_P$ and $\eta_E$, their difference is bounded by the corresponding exact-arithmetic bound plus $\eta_P+\eta_E$.

\begingroup
\setlength{\abovedisplayskip}{5pt}
\setlength{\belowdisplayskip}{5pt}
\setlength{\textfloatsep}{8pt plus 2pt minus 2pt}
\section{Proxy Rank and Threshold Analysis}
\label{app:rank_diagnostics}

We investigate the effects of proxy rank and refresh threshold from
two complementary perspectives: approximation accuracy under fixed
queries and the resulting reasoning and routing behavior under
configuration-specific decoding trajectories.

\smallskip
\noindent\textbf{Fixed-query approximation analysis.}
We first revisit the 32 MathVista trajectories generated by Qwen3-VL-8B
in Sec.~\ref{sec:ablation_rank}, using the default configuration
$r=16$, $\theta_M=0.10$, and the cosine cache-age schedule with $D=512$.
With queries and visual keys fixed, we evaluate all six ranks every
eight decoding steps until step 512 or natural termination.
The 1,167 sampled positions across 32 trajectories yield 1,344,384
paired head--step observations per rank (36 layers, 32 query heads).
We use the exact BF16 visual attention mass as the reference, retaining
finite-precision effects even at the largest rank. This controlled
comparison isolates the effect of approximation rank from differences
in decoding trajectories.

\smallskip
\noindent\textbf{Rank--threshold evaluation.}
To examine the interaction between approximation rank and refresh
policy, we evaluate all 24 combinations of
$r\in\{4,8,16,32,64,128\}$ and
$\theta_M\in\{0.05,0.10,0.15,0.20\}$.
For all configurations, we set $D=512$ and use the indicated fixed
base threshold across inputs, disabling input-dependent prefill
threshold calibration. The operative threshold still follows the
cosine cache-age schedule.
We randomly sample 512 examples from the 1,000-question MathVista
testmini split and evaluate all configurations on this shared subset.
This provides a controlled sensitivity analysis, rather than an
independent held-out parameter-selection result.
For a more detailed analysis of routing behavior, we randomly select
16 examples from the same 512-example subset using a fixed seed.
Each configuration generates its own greedy decoding trajectories
under this fixed-base-threshold protocol. Across the resulting 384
configuration--example runs, we record proxy and exact visual mass
at every eighth decoding step, up to step 512 or natural termination.
Measurements are restricted to Persist checks, including those that
trigger an immediate Refresh. Each configuration is evaluated along
its own trajectory, using its corresponding queries, cache ages, and
age-adjusted thresholds. Heads already in Refresh and mandatory
cache-age refreshes are excluded, as neither involves a proxy-based
routing decision.

\begin{figure}[!t]
\newcounter{vaporalgo}
\refstepcounter{vaporalgo}\label{alg:vapor}
\hrule\smallskip
\textbf{Algorithm \thevaporalgo: VAPOR attention procedure}
\smallskip\hrule\smallskip
\begingroup
\footnotesize
\setlength{\tabcolsep}{0pt}
\renewcommand{\arraystretch}{0.94}
\newcommand{\vaporindent}{\hspace*{0.8em}}
\begin{tabular}{@{}r@{\hspace{0.6em}}p{\dimexpr\columnwidth-2.1em\relax}@{}}
 & \textbf{Input:} visual/text KV cache, rank $r$, maximum age $D$, threshold mapping $f$. \\
 & \textbf{Preparation after prefill} \\
1 & Compute $C_h$ for valid prompt-token distributions. \\
2 & $C\gets\operatorname{median}_h C_h$, $\theta_M\gets f(C)$. \\
3 & For each distinct visual key matrix, compute \\
  & \vaporindent $K_V\approx U_r\Sigma_rV_r^{\mathsf T}$. \\
4 & Initialize $\mathrm{state}_h\gets\mathrm{Refresh}$ for all $h$. \\
 & \textbf{Decoding attention} \\
5 & \textbf{for} each decoding step $t=1,2,\ldots$ \textbf{do} \\
6 & \vaporindent \textbf{for} each head $h$ in model execution order \textbf{do} \\
7 & \vaporindent\vaporindent Obtain $q_t$ and current text keys and values. \\
8 & \vaporindent\vaporindent Compute exact text logits, $Z_{T,t}$, and $\bar u_t$. \\
9 & \vaporindent\vaporindent $b\gets(\mathrm{state}_h=\mathrm{Refresh})$. \\
10 & \vaporindent\vaporindent \textbf{if} $\mathrm{state}_h=\mathrm{Persist}$ \textbf{then} \\
11 & \vaporindent\vaporindent\vaporindent $\delta\gets t-a_h$. \\
12 & \vaporindent\vaporindent\vaporindent \textbf{if} $\delta\geq D$ \textbf{then} $b\gets\mathrm{true}$. \\
13 & \vaporindent\vaporindent\vaporindent \textbf{else} \\
14 & \vaporindent\vaporindent\vaporindent\vaporindent Compute $\tau_{h,t}$ using Eq.~\ref{eq:age_threshold}. \\
15 & \vaporindent\vaporindent\vaporindent\vaporindent $\hat z_{V,t}\gets U_r\Sigma_r(V_r^{\mathsf T}q_t)/\sqrt{d_h}$. \\
16 & \vaporindent\vaporindent\vaporindent\vaporindent Compute $M_{P,t}$ using Eq.~\ref{eq:proxy_mass}. \\
17 & \vaporindent\vaporindent\vaporindent\vaporindent $b\gets(M_{P,t}\geq\tau_{h,t})$. \\
18 & \vaporindent\vaporindent\vaporindent \textbf{end if} \\
19 & \vaporindent\vaporindent \textbf{end if} \\
20 & \vaporindent\vaporindent \textbf{if} $b$ \textbf{then} \hfill $\triangleright$ Refresh now \\
21 & \vaporindent\vaporindent\vaporindent Compute exact $z_{V,t}$, $M_{E,t}$, and $\bar v_t$. \\
22 & \vaporindent\vaporindent\vaporindent $o_t\gets M_{E,t}\bar v_t+(1-M_{E,t})\bar u_t$. \\
23 & \vaporindent\vaporindent\vaporindent Cache $\bar v_{a_h}\gets\bar v_t$ and set $a_h\gets t$. \\
24 & \vaporindent\vaporindent\vaporindent $\mathrm{state}_h\gets\mathrm{Persist}$ if $M_{E,t}<\theta_M$, \\
   & \vaporindent\vaporindent\vaporindent otherwise $\mathrm{state}_h\gets\mathrm{Refresh}$. \\
25 & \vaporindent\vaporindent \textbf{else} \hfill $\triangleright$ Persist \\
26 & \vaporindent\vaporindent\vaporindent $o_t\gets M_{P,t}\bar v_{a_h}+(1-M_{P,t})\bar u_t$. \\
27 & \vaporindent\vaporindent\vaporindent Keep $\mathrm{state}_h=\mathrm{Persist}$ and the anchor. \\
28 & \vaporindent\vaporindent \textbf{end if} \\
29 & \vaporindent\vaporindent Collect $o_t$ and, once all heads in this layer finish, concatenate their outputs and apply the standard output projection. \\
30 & \vaporindent \textbf{end for} \\
31 & \textbf{end for} \\
\end{tabular}
\endgroup
\smallskip\hrule
\end{figure}

\smallskip
\noindent\textbf{Evaluation metrics.}
We assess each configuration in terms of reasoning accuracy,
visual-mass approximation error, and refresh-decision agreement.
For $N$ recorded head--step pairs, let $\tau_i$ denote the effective
age-adjusted threshold. The proxy-based and exact-mass-based refresh
decisions are defined as
\[
    \hat d_i=\mathbf{1}[M_{P,i}\geq\tau_i],\qquad
    d_i=\mathbf{1}[M_{E,i}\geq\tau_i].
\]
We compute mass mean absolute error (MAE) and decision agreement as
\[
\begin{aligned}
    \mathrm{MAE}
    &= \frac{100}{N}\sum_i|M_{P,i}-M_{E,i}|,\\
    A
    &= \frac{1}{N}\sum_i\mathbf{1}[\hat d_i=d_i].
\end{aligned}
\]
Mass MAE is expressed in percentage points, while accuracy and
decision agreement are reported as percentages. Both metrics are
pooled over head--step pairs with equal weight.

\smallskip
\noindent\textbf{Results and discussion.}
Table~\ref{tab:rank_threshold_grid} reveals a consistent relationship
between proxy rank and approximation quality: increasing the rank
reduces mass MAE and improves decision agreement at every tested
threshold. For example, at $\theta_M=0.10$, increasing the rank from
4 to 128 reduces MAE from 2.430 to 0.020 percentage points and
improves decision agreement from 89.18\% to 99.88\%.
However, these improvements in approximation and routing consistency
do not translate into monotonically increasing reasoning accuracy.
The highest observed accuracy is 77.54\%, achieved at
$r=16$ and $\theta_M=0.10$, while larger ranks provide no further
accuracy gains on this randomly sampled subset.
Among the four tested thresholds, the highest observed accuracy occurs
at $\theta_M=0.05$ for $r\in\{4,8\}$, $0.10$ for $r\in\{16,32\}$,
and $0.15$ for $r\in\{64,128\}$.
Smaller thresholds can compensate for the systematic visual-mass
underestimation at lower ranks observed in
Fig.~\ref{fig:rank_mass_density}, allowing underestimated proxy masses
to trigger Refresh sooner. These results suggest that maximizing proxy
fidelity is not necessarily optimal for downstream reasoning.
Instead, proxy rank should balance estimation accuracy against the
computational cost of proxy evaluation and its impact on refresh
decisions. In particular, $r=16$ achieves the highest observed
accuracy among the tested configurations while avoiding the additional
rank-dependent computation required by larger approximations.
We therefore adopt $r=16$ as the default in our main experiments.

\endgroup
\section{VAPOR Algorithm}
\label{app:algorithm}

Algorithm~\ref{alg:vapor} summarizes Sec.~\ref{sec:persistence_framework}.
The index $h$ denotes a layer--query-head pair, with positional encoding included in queries and keys.
Heads sharing visual keys share low-rank factors, but retain separate states, anchor steps, and cached conditional visual outputs.
Textual KV caches update once per layer before head evaluation.
The base threshold $\theta_M$ is calibrated per input using the rule in Sec.~\ref{sec:setup}.

\end{document}